\documentclass{article} 
\usepackage[preprint]{colm2025_conference}

\usepackage{microtype}
\usepackage{hyperref}
\usepackage{url}
\usepackage{booktabs}

\usepackage{lineno}

\usepackage{wrapfig}
\usepackage[inline]{enumitem}

\usepackage{multirow,mathtools } 
\usepackage{algorithm,algpseudocode}

\usepackage{pifont}
\usepackage{color, colortbl}
\usepackage{graphicx}
\usepackage{blindtext}
\usepackage{lipsum}

\usepackage{threeparttable}
\usepackage{adjustbox}

\usepackage{amsmath}  
\usepackage{amsfonts}  
\usepackage{todonotes}
\usepackage[most]{tcolorbox} 
\definecolor{LightCyan}{rgb}{0.88,1,1}
\definecolor{Gray}{gray}{0.93}
\definecolor{darkblue}{rgb}{0, 0, 0.5}

\hypersetup{colorlinks=true, citecolor=darkblue, linkcolor=darkblue, urlcolor=darkblue}

\newcommand{\SL}[1]{\textcolor{red}{SL: #1}}
\newcommand{\SP}[1]{\textcolor{blue}{SP: #1}}

\newcommand{\Df}{\mathcal D_\mathrm{f}}

\newcommand{\Dr}{\mathcal D_\mathrm{r}}
\newcommand{\lf}{\ell_\mathrm{f}}
\newcommand{\lr}{\ell_\mathrm{r}}
\newcommand{\btheta}{{\boldsymbol{\theta}}}

\newcommand{\thatis}{{\textit{i.e.}}}

\newcommand{\grand}{\textsc{GraNd}}
\newcommand{\mink}{\textsc{Min-K\% Prob}}
\newcommand{\moderate}{\textsc{Moderate}}
\newcommand{\random}{\textsc{Random}}

\newcommand{\reffig}[1]{\textbf{Fig.\,\ref{#1}}}
\newcommand{\reftab}[1]{\textbf{Table\,\ref{#1}}}

\newtcolorbox{prompt}[1]{
    enhanced,
    drop shadow=black!5!white,
    left=4mm,
    right=4mm,
    top=2mm,
    bottom=2mm,
    boxsep=0mm,
    rounded corners,
    title=#1,
    fontupper=\footnotesize\linespread{0.9}\fontfamily{lmr}\selectfont,
    }

\everydisplay{\small}
\newcommand{\Def}[0]{\mathrel{\mathop:}=}

\title{LLM Unlearning  Reveals a Stronger-Than-Expected Coreset Effect in Current Benchmarks}

\author{Soumyadeep Pal$^{\dag, \star}$~~~~Changsheng Wang$^{\dag, \star}$~~~~
\textbf{James Diffenderfer}$^\ddag$\\
\textbf{Bhavya Kailkhura}$^\ddag$~~~~\textbf{Sijia Liu}$^{\dag, \S}$  \\
$^\dag$Michigan State University, $^\ddag$Lawrence Livermore National Laboratory, $^\S$IBM Research\\
$^\star$Equal contribution
}

\begin{document}

\ifcolmsubmission
\linenumbers
\fi

\maketitle





\begin{abstract}
Large language model (LLM) unlearning has become a critical challenge in ensuring safety and controlled model behavior by removing \textit{undesired} data-model influences from the pretrained model while preserving its general utility. 
Significant recent efforts have been dedicated to developing LLM unlearning benchmarks such as WMDP (Weapons of Mass Destruction Proxy) and MUSE (Machine Unlearning Six-way Evaluation), facilitating standardized unlearning performance assessment and method comparison. Despite their usefulness, we uncover for the first time a novel \textit{coreset effect} within these benchmarks. Specifically, we find that LLM unlearning achieved with the original (full) forget set can be effectively maintained using a significantly smaller subset (functioning as a ``coreset''), \textit{e.g.}, as little as 5\% of the forget set, even when selected at random.
This suggests that LLM unlearning in these benchmarks can be performed surprisingly easily, even in an extremely low-data regime.
We demonstrate that this coreset effect remains strong, regardless of the LLM unlearning method used, such as NPO (Negative Preference Optimization) and RMU (Representation Misdirection Unlearning), the popular ones in these benchmarks. 
The surprisingly strong coreset effect is also robust across various data selection methods, ranging from random selection to more   sophisticated heuristic  approaches.
We explain the coreset effect in LLM unlearning  through a keyword-based perspective, showing that keywords extracted from the forget set alone contribute significantly to unlearning effectiveness and indicating that current unlearning is driven by a compact set of high-impact tokens rather than the entire dataset. We further justify the faithfulness of coreset-unlearned models 
along additional dimensions, such as mode connectivity and robustness to jailbreaking attacks. Codes are available at \href{https://github.com/OPTML-Group/MU-Coreset}{https://github.com/OPTML-Group/MU-Coreset}.
\end{abstract}

\vspace*{-3mm}
\section{Introduction}
\label{sec: intro}
\vspace*{-3mm}

The problem of machine unlearning (MU) for large language models (LLMs), referred to as LLM unlearning \citep{liu2025rethinking,si2023knowledge,qu2024frontier,cooper2024machine}, is gaining critical importance as a means of enforcing data privacy rights (\textit{e.g.}, preventing the generation of copyrighted or sensitive content) \citep{eldan2023whos,shi2024muse,maini2024tofu,jang2022knowledge}, and for removing harmful or unsafe knowledge embedded in models amid growing concerns around safety and alignment \citep{li2024wmdp,yao2024large,barez2025open,zhang2024safe,chen2025safety}.
The core objective of MU is to remove the influence of specific, undesired data or knowledge from a trained model, while preserving its general utility, without the cost of full retraining from scratch.

Despite the growing importance of LLM unlearning, much of the existing research has primarily focused on the design of unlearning \textit{algorithms}. Notable approaches include gradient ascent and its variants \citep{thudi2022unrolling,jang2022knowledge,yao2024large}, which aim to reverse the training effect of the forget data by explicitly pushing the model away from learned patterns; influence function-based methods \citep{izzo2021approximate,jia2024soul,koh2017understanding}, which assess and mitigate the contribution of individual data points to model behavior; preference optimization techniques and their extensions \citep{rafailov2023direct,zhang2024negative,fan2024simplicity}, which guide the model to reduce its preference for responses associated with the forget data; misrepresentation learning methods \citep{li2024wmdp}, which disrupt internal representations linked to the undesired knowledge; and neuron- or model-editing approaches that leverage task vectors or localized model components to guide unlearning \citep{jia2024wagle,hase2023does,wu2023depn}.

While the above LLM unlearning algorithms have advanced the field, the \textit{data perspective} of LLM unlearning, such as how the composition or size of the forget set (\textit{i.e.}, the dataset used to define the unlearning scope) influences unlearning performance, has received significantly less attention.
On the data side, some recent efforts  explored evaluating the memorization level of forget data to better guide unlearning optimization  \citep{barbulescu2024each,zhao2024makes}.
Yet, most existing efforts focused on developing LLM unlearning \textit{benchmark datasets}, such as TOFU (Task of
Fictitious Unlearning) to remove synthesized fictitious information \citep{maini2024tofu},
{WMDP} (Weapons of Mass Destruction Proxy)  for harmful knowledge removal \citep{li2024wmdp} and  {MUSE} (Machine Unlearning Six-way Evaluation) for copyrighted data removal \citep{shi2024muse} and other similar datasets \citep{eldan2024whos,liu2024revisiting}. 

Although the above benchmarks have served as standardized platforms for LLM unlearning training and evaluation, 
the forget datasets curated in these benchmarks are typically provided as fixed, full sets, and existing research has largely adopted them without questioning the appropriateness, sufficiency, or redundancy of these forget sets for achieving effective unlearning.
For example, in WMDP, a large forget data source is provided containing sensitive documents (25K articles) related to hazardous knowledge in biosecurity and cybersecurity. However, the forget set actually used in the WMDP-based unlearning via RMU (Representation Misdirection Unlearning) utilizes only the first 600 articles from these source documents \citep{li2024wmdp}, leaving open the question of whether the full dataset is necessary or if a much smaller subset could suffice.
Therefore, in this work, we ask:
\begin{tcolorbox}[before skip=2mm, after skip=0.0cm, boxsep=0.0cm, middle=0.0cm, top=0.05cm, bottom=0.05cm, boxrule=0.6pt]
\begin{center}
     \textit{\textbf{(Q)} Does there exist a ``coreset'' within the LLM unlearning forget dataset that can yield lossless unlearning performance?}
\end{center}
\end{tcolorbox} 
\vspace*{2mm}

The term ``coreset'' is inspired by prior research on coreset selection
for non-LLMs \citep{guo2022deepcore, borsos2020coresets, zhang2023selectivity}, where a subset of the 
training set 
enables comparable model performance to training on the full dataset.
We envision that
addressing (Q) could open up a new, underexplored paradigm: LLM unlearning in the low-data regime, where effective unlearning is achieved using only a small subset of the forget data.
It may also reveal potential weaknesses in current unlearning benchmarks \citep{thaker2024position}, where unlearning could be surprisingly easy due to redundancy in the forget set.

In this work, we provide an affirmative answer to (Q) through extensive empirical studies: a coreset effect does exist in LLM unlearning, and unlearning in current benchmarks such as WMDP and MUSE can be effectively achieved using as little as 5\% of the original full forget set, even when the coreset is randomly selected. We summarize \textbf{our contributions} below.

$\bullet$ We unveil a novel \textit{Coreset Effect} in LLM unlearning, showing that popular unlearning methods (RMU and negative preference optimization) achieve comparable performance on benchmarks like WMDP and MUSE using as little as 5\% of the forget set. 

$\bullet$ We demonstrate that this effect holds across both random and more sophisticated coreset selection strategies. We also provide a keyword-based analysis to explain the strong coreset effect where small forget subsets can drive unlearning behavior.

$\bullet$ We show that coreset-unlearned models exhibit similar unlearning characteristics to full-set-unlearned ones across dimensions including mode connectivity, adversarial robustness, and model utility, except for a potential increased vulnerability to relearning attacks.

\vspace*{-3mm}
\section{Related Works}
\label{sec: works}
\vspace*{-3mm}

\noindent \textbf{LLM Unlearning.}
Unlearning seeks to remove the influence of undesired data-model influence to protect privacy or prevent harmful knowledge generation. A commonly-used gold standard is exact unlearning--retraining the model from scratch without the forget data (named ``Retrain'') \citep{cao2015towards,thudi2022unrolling} --but this is computationally prohibitive, especially for LLMs. As a result, significant efforts have focused on \textit{approximate} unlearning \citep{bourtoule2021machine, liu2025rethinking}, which modifies pretrained model weights post hoc. These approaches fall broadly into model finetuning methods \citep{ilharco2022editing, li2024wmdp, zhang2024negative, fan2024simplicity, jia2024wagle}, and input-based prompting methods \citep{pawelczyk2023context, thaker2024guardrail}. Recent studies have also highlighted vulnerabilities in these methods, including jailbreaking \citep{lucki2024adversarial, lynch2024eight}, latent extraction \citep{seyitouglu2024extracting}, and fine-tuning-based relearning attacks \citep{hu2024jogging, deeb2024unlearning}, revealing fragility in unlearning robustness. Despite this progress, \cite{thaker2024position} argues that current benchmarks offer an \textit{overly optimistic view} due to weak evaluation protocols. 
As a complementary concern, we uncover a strong coreset effect in standard LLM unlearning benchmarks, showing that effective unlearning can be achieved using surprisingly small subsets of the forget data, which raises potential questions about the reliability and rigor of current unlearning dataset designs.

\noindent \textbf{Data perspective in unlearning.}
Recent work has begun to explore the influence of forget data on unlearning outcomes, particularly through the lens of data ordering and importance. For instance, \cite{zhao2024makes} introduced RUM, a meta-unlearning algorithm that partitions the forget set into homogeneous subsets using data quality scores, such as representation distance from the centroid and memorization score, and applies specialized unlearning strategies to each subset
sequentially.
\cite{zhao2024scalability} extended this framework by evaluating additional memorization proxies for sample partitioning.
Both of these studies are limited to image classification tasks. In the context of LLMs, \cite{barbulescu2024each} proposed a dynamic unlearning approach that iteratively selects high-memorization samples from the forget set in each epoch. While these efforts highlight the role of data quality or ordering in unlearning, they do not examine the existence of a {coreset}.


\noindent \textbf{Coreset selection.}
Coreset selection, developed mainly for classification using computer vision models,
refers to the process of selecting a subset of the training data, which can maximize the model accuracy on said task when trained using such subset. One class of coreset selection methods typically assign importance scores to each sample in the dataset based on various metrics computed during training dynamics such as gradient norm \citep{paul2021deep}, l2 norm of error vectors \citep{paul2021deep}, forgetting score \citep{toneva2018an}, data loss \citep{welling2009herding,pruthi2020estimating}  and prediction confidence \citep{pleiss2020identifying}. Additionally some methods also focus on data diversity, based on distances of datapoints from their class clusters \citep{xia2022moderate} and coverage of the training set \citep{zheng2022coverage, maharana2024mathbbd}. Another class of methods are optimization based approaches for coreset selection mainly focusing on gradient information of data samples \citep{mirzasoleiman2020coresets,killamsetty2021grad,killamsetty2021glister,pooladzandi2022adaptive}.

For language models, alongside loss-based scoring \citep{azeemi2023data}, various classic scoring-based coreset selection methods (mentioned above) have been explored for downstream finetuning tasks where a smaller model is used for efficient scoring \citep{zhang2025staff}. Additionally, coreset selection methods have been developed for instruction-tuning where the influence of each training sample is computed using gradient similarity with samples in the validation set \citep{xia2024less} and using LLM-as-a-Judge \citep{chen2024alpagasus,liu2023makes}. Finally, \cite{zhou2023lima} demonstrated superior performance LLM-alignment by finetuning on an extremely small set of manually curated high quality dataset.

To the best of our knowledge, \cite{patil2025upcore} {concurrently} explores coreset selection in LLM unlearning. Using anomaly detection on the hidden state representations of the forget set, the authors prune about 10-30\% of the forget set to determine the coreset with the primary goal of utility preservation. However, our work differs from this work in two key aspects: 
(1) Their study did not identify the strong coreset effect embedded in existing LLM unlearning benchmarks, nor did it characterize its conditions (\textit{e.g.}, the relationship between coreset effect and unlearning training time); And (2) we go further by investigating both the ``what'' and ``why'' behind the coreset effect, focusing on extreme pruning ratios (90-99\%), \textit{i.e.}, extreme low-data regimes  (1–10\%),   while preserving unlearning and utility performance, and analyzing its impact on unlearning robustness.



\vspace*{-3mm}
\section{Preliminaries and Motivation on Coreset Effect in LLM Unlearning}
\label{sec: problemstmnt}
\vspace*{-3mm}

\noindent \textbf{LLM unlearning setup.}
LLM unlearning aims to remove undesired data, knowledge, or model influence from a pretrained LLM, while preserving its general utility. This is typically framed as a model fine-tuning or editing task, given access to a forget set $\Df$, which contains the data or knowledge to be unlearned, and a retain set $\Dr$, which is typically unrelated to the unlearning target and serves to help preserve the model’s overall utility post-unlearning.
Therefore, the problem of LLM unlearning is cast as the following regularized optimization framework \citep{liu2025rethinking,yao2024large,maini2024tofu},
\begin{align}
\begin{array}{l}
 \displaystyle \min_{\boldsymbol{\theta}}
 \,\, {\mathbb E_{(x, y) \in \Df} [ \lf(y | x; \boldsymbol{\theta}) ]} + 
 \lambda { \mathbb E_{(x, y) \in \Dr} [ \lr( y | x; \boldsymbol{\theta}) ] },
\end{array}
\label{eq: LLM_MU}
\end{align} 
where $\boldsymbol{\theta}$ denotes the model parameters to be updated during unlearning (initialized from the pretrained model state), $\mathcal{L}_{\mathrm{f}}$ and $\mathcal{L}_{\mathrm{r}}$ denote the forget loss (\textit{i.e.}, the unlearning objective) and retain loss (\textit{i.e.}, the utility-preserving objective), respectively, evaluated when using $\boldsymbol{\theta}$ to generate a response $y$ given input $x$, and $\lambda >0$ is a regularization parameter to strike the balance between the forget and retain objectives.

 \noindent 
\textbf{Representative LLM unlearning methods considered in this study.}
The specifics of the unlearning optimization in \eqref{eq: LLM_MU} correspond to different LLM unlearning approaches, typically depending on how the forget loss $\lf
$ is formulated and implemented.
Although numerous unlearning methods have been proposed in the literature, our work focuses on two widely used and high-performing LLM unlearning approaches, {negative preference optimization} (\textbf{NPO}) \citep{zhang2024negative} and 
 {representation misdirection unlearning} (\textbf{RMU}) \citep{li2024wmdp}, 
 where NPO typically excels at data-wise unlearning, as demonstrated on the MUSE benchmark \citep{shi2024muse}, and RMU is particularly effective for knowledge unlearning, as evidenced by its strong performance on the WMDP benchmark \citep{li2024wmdp}.
NPO is adapted from direct preference optimization (DPO) \citep{rafailov2023direct}, but differs by treating the forget data in $\Df$ as negative examples, excluding any positive examples. This yields a preference-based forget loss $\lf$  that drives the unlearned model $\btheta$ to deviate from the original, pre-trained model (referred to as the reference model in NPO) on the forget set $\Df$.  The retain loss $\lr$ in NPO is typically chosen as the standard cross-entropy next-token prediction loss on the retain set $\Dr$.
 In contrast to NPO, RMU is built upon a random feature-based unlearning objective \citep{li2024wmdp,fan2024challenging}. The forget loss $\lf$ in RMU promotes unlearning by mapping the intermediate-layer representations of the forget data to random features, thereby disrupting the model’s ability to generate or reconstruct the forgotten information. 
In a similar spirit, the retain loss $\lr$ enforces alignment between the representations of the unlearned model and the pre-unlearned (reference) model on the retain dataset, helping to preserve the model’s general utility.
We refer readers to Appendix\,\ref{app: npormu} for the detailed formulations of NPO and RMU.

\noindent 
\textbf{Unlearning benchmarks considered in this study.} 
We conduct and evaluate LLM unlearning using two benchmarks: \textbf{WMDP} \citep{li2024wmdp}, which assesses the unlearning of potentially hazardous knowledge in domains such as 
biology (called \textbf{WMDP-Bio}) and cybersecurity (called \textbf{WMDP-Cyber});
and \textbf{MUSE} \citep{shi2024muse}, which focuses on unlearning textual content from Harry Potter books (named \textbf{MUSE-Books})
and news articles (named \textbf{MUSE-News}).
We select these two benchmarks as the former is the representative \textit{knowledge} unlearning benchmark and the latter is designed to additionally support \textit{data} unlearning (\textit{e.g.}, to prevent verbatim memorization  and privacy leakage). It is worth noting that other unlearning benchmarks also exist, such as TOFU \citep{maini2024tofu}. However, the evaluation of unlearning effectiveness in TOFU relies on a $p$-value metric, which can be insensitive in distinguishing unlearning performance when the $p$-value exceeds $0.1$, making it less effective for fine-grained comparisons between unlearning methods. 

In WMDP \citep{li2024wmdp}, the goal of unlearning LLM is to reduce its \textit{accuracy on the WMDP evaluation set}, 
which comprises QA pairs with undesirable answers related to harmful knowledge targeted for removal.
Accordingly, we measure \textbf{unlearning effectiveness} (\textbf{UE}) as:
{$\text{UE} = 100\% - (\text{Accuracy on WMDP evaluation set})$}. 
Since the associated QA pairs are multiple-choice questions with 4 choices, a perfect unlearning process would achieve an UE of 75\%.
The \textbf{utility} (\textbf{UT}) of the WMDP-unlearned model is measured by the zero-shot accuracy on the \textit{MMLU} dataset \citep{hendrycks2020measuring}. 

In MUSE \citep{shi2024muse}, \textbf{UE} is measured using different metrics: (1) \textit{Verbatim memorization} (\textbf{VerbMem}) on the forget set $\Df$ reflects the model’s ability to perform next-token prediction for completing the forgotten data records. 
(2) \textit{Knowledge memorization} (\textbf{KnowMem}) reflects the model's ability to answer questions involving  undesired knowledge in MUSE. Thus,
a lower VerbMem (or KnowMem) indicates better UE, as it implies reduced model generation capability for the targeted data (or knowledge) removal.
Besides VerbMem and KnowMem, UE in MUSE is also evaluated using (3) \textit{privacy leakage} (\textbf{PrivLeak}), which assesses the extent to which the unlearned model leaks membership information, \textit{i.e.}, whether it reveals that data in $\Df$ was part of the original training set.  PrivLeak values approaching zero indicate better unlearning. 
\textbf{UT} of the unlearned model is measured by  \textbf{KnowMem on MUSE's retain set $\Dr$}, reflecting the model’s ability to preserve useful knowledge unrelated to   unlearning.

Furthermore, we note that when implementing NPO and RMU to solve problem \eqref{eq: LLM_MU} for WMDP and MUSE, the number of training epochs is typically kept \textit{small}. This is because the forget loss encourages divergence from the pretrained model, and prolonged training may lead to model collapse or degradation of general utility. For example, \textit{the number of unlearning epochs for RMU is set to 1} for WMDP and MUSE-Books. We refer the readers to Appendix\,\ref{app: epochs} for   a detailed discussion of the unlearning settings across different benchmarks.





\noindent 
\textbf{Problem of interest: ``Coreset'' effect in LLM unlearning  under low-data regimes.}
In current unlearning setups, a key {pre-condition} is the assumption of access to the forget set $\Df$, which is typically provided \textit{a priori} in current benchmarks. However, very few studies have investigated the problem of LLM unlearning in a {low-data} regime, \textit{i.e.}, when the size of the forget set $\Df$ is limited or small. Therefore, the primary problem of interest in this work is to explore whether a ``coreset''-like effect exists in LLM unlearning, thereby enabling effective unlearning  in low-data regimes of current benchmarks.

\begin{figure}[htb]
\vspace*{-2.8mm}
\centering
\begin{tabular}{cc} \includegraphics[width=.5\textwidth]{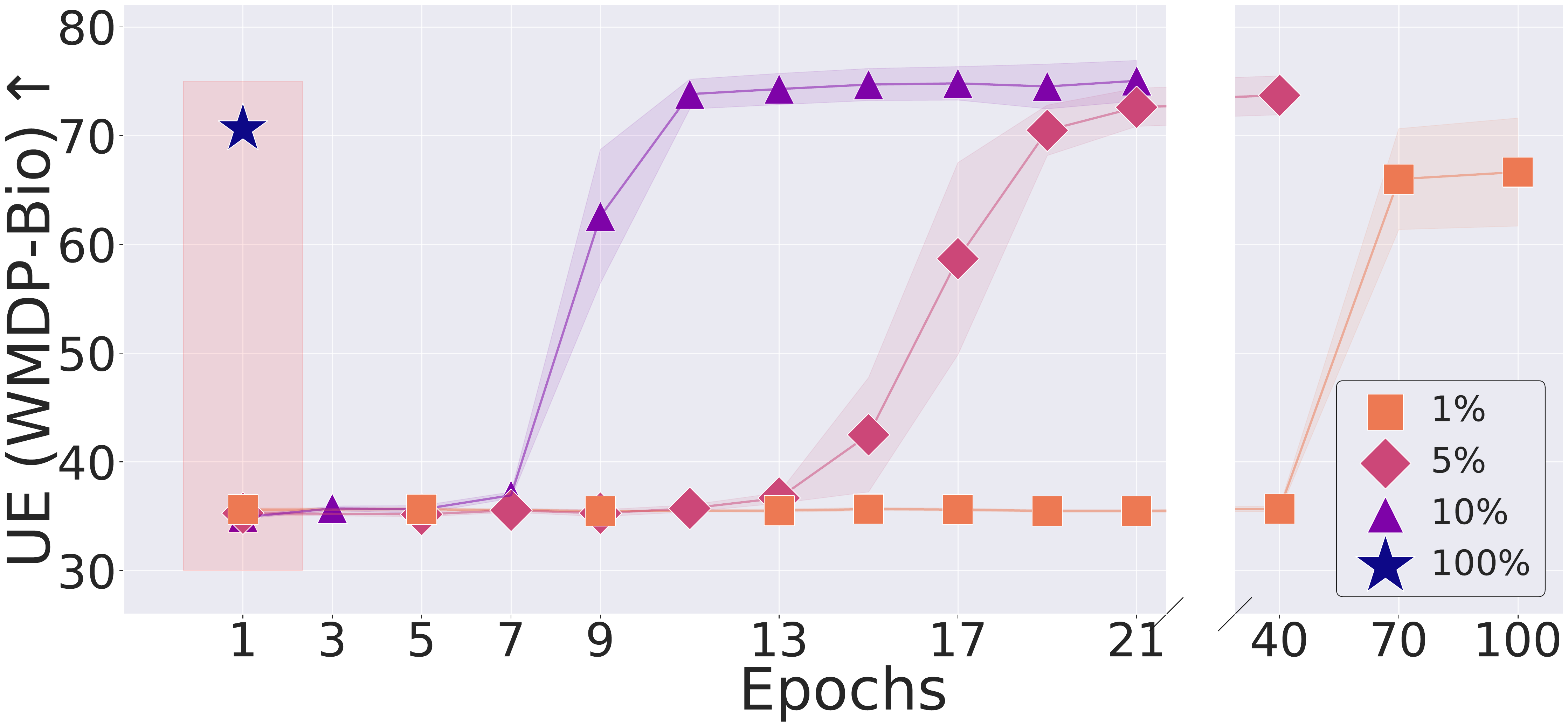}    &  \includegraphics[width=.32\textwidth]{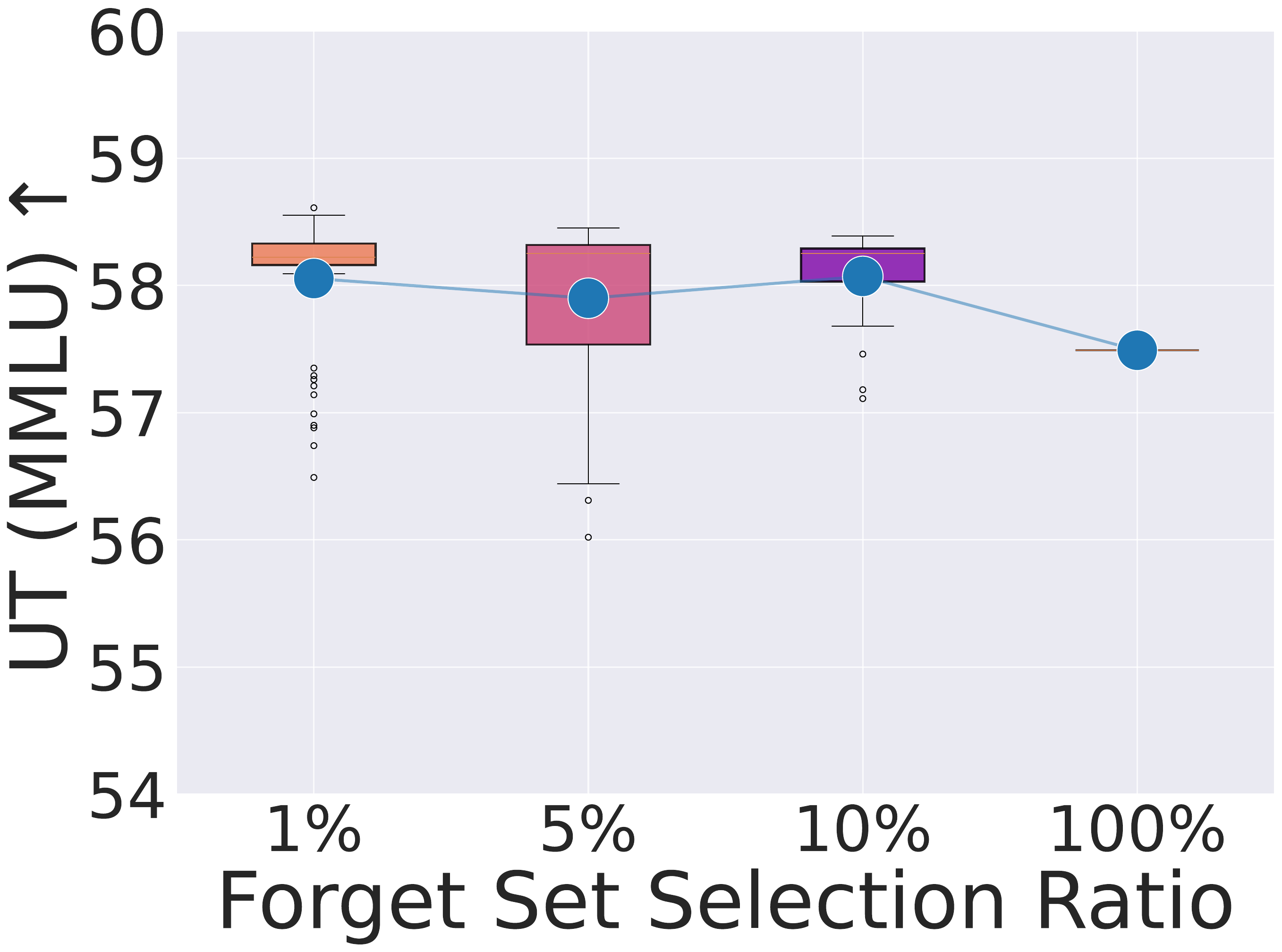}  \\
  {\footnotesize{(a) UE vs unlearning epoch \#}}    & \hspace*{2mm} {\footnotesize{(b) UT vs. coreset size}} 
\end{tabular}
\vspace*{-3mm}
\caption{\small{
Unveiling the ``coreset''-like effect in LLM unlearning on WMDP using the RMU method, applied to the pre-trained LLM 
Zephyr-7B-$\beta$. The  ``coreset'' is randomly sampled from the full forget set ($\Df =$ WMDP-Bio), with selection ratios of 1\%, 5\%, and 10\%; the 100\% setting corresponds to using the entire $\Df$. Unlearning performance is averaged over 5 random trials. 
(a) The ``coreset'' achieves comparable UE (unlearning effectiveness) for RMU  on WMDP-Bio, especially when unlearning is performed with longer training epochs. The default number of unlearning epochs is 1 for RMU under the full $\Df$ (100\%), as indicated by the shaded region.
(b) MMLU-based UT (utility) of the unlearned model against the forget set selection ratio. Each box plot represents the UT performance of the unlearned model across the range of unlearning epochs shown in (a) for 5  random trials.
%
}}
  \label{fig: coreset-motivation}
  \vspace*{-2mm}
\end{figure}

Through the motivating results in \textbf{Fig.\,\ref{fig: coreset-motivation}}, we demonstrate the existence of a coreset effect--where a small random subset of the original forget set has been able to achieve comparable test-time unlearning performance, \textit{especially when unlearning is trained for extended epochs}.
To be specific,
 Fig.\,\ref{fig: coreset-motivation} presents the unlearning performance in terms of both UE (unlearning effectiveness) and UT (utility measured on MMLU), using the RMU unlearning method applied to the pre-trained LLM 
Zephyr-7B-$\beta$ on the WMDP benchmark. The performance is evaluated across different random data selection ratios (1\%, 5\%, and 10\%) with respect to the full forget set WMDP-Bio.
Fig.\,\ref{fig: coreset-motivation}(a) shows that an unlearned model trained on a small subset of the forget set (\textit{e.g.}, as little as 5\% of the full set) can achieve UE almost the same as that of using the entire forget set (100\%). However, \textbf{this coreset effect only emerges when the training is extended beyond the RMU's default 1-epoch setting on WMDP}; For example, using only 1\% of the forget set requires over 70 training epochs to achieve comparable UE. 
In addition, {the observed effectiveness of coreset unlearning does \textbf{not} come at the cost of UT}. This is evidenced by Fig.\,\ref{fig: coreset-motivation}(b), where the unlearning optimization (via RMU) maintains UT on MMLU across varying unlearning epochs (indicated by the box regions) and coreset ratios, in comparison to the model unlearned using the 100\% forget set.

The above motivating results provide strong evidence for a coreset effect in LLM unlearning, showing that with sufficient training, small randomly selected subsets can achieve comparable UE and UT to full forget set-based unlearning. This finding prompts several key questions: \textit{Does the effect hold  or even improve with non-random coreset selection?} \textit{What underlying factors drive this phenomenon?} And \textit{are coreset-unlearned models as faithful and robust as their full-set counterparts?} We investigate these questions in the following sections.

\vspace*{-3mm}
\section{On the Consistency and Rationale Behind Forget Coreset Effect}
\label{sec: consistency}
\vspace*{-3mm}

In this section, we present a comprehensive study of the coreset effect in LLM unlearning across various dimensions, including different unlearning approaches (RMU and NPO), benchmarks (WMDP and MUSE), and coreset selection methods. Furthermore, we explore a possible explanation for this effect through the lens of keywords extracted from the coreset, which may serve as a key driving force behind effective unlearning.

\noindent 
\textbf{Surprisingly strong and consistent performance of \textit{random} coreset selection.}
Recall from our motivating results on WMDP-Bio unlearning using RMU in Fig.\,\ref{fig: coreset-motivation}, that coresets formed via random data selection already demonstrate lossless UE while maintaining UT.
To further validate the effectiveness of random selection (termed \random{}), we investigate another two key aspects: (1) random coreset unlearning across different unlearning methods and benchmarks, and (2) comparison with other more sophisticated coreset selection methods.

\begin{figure}[htb]
\vspace*{-2mm}
\centering
\begin{tabular}{cccc}
  \hspace*{-1mm} \includegraphics[width=.225\textwidth]{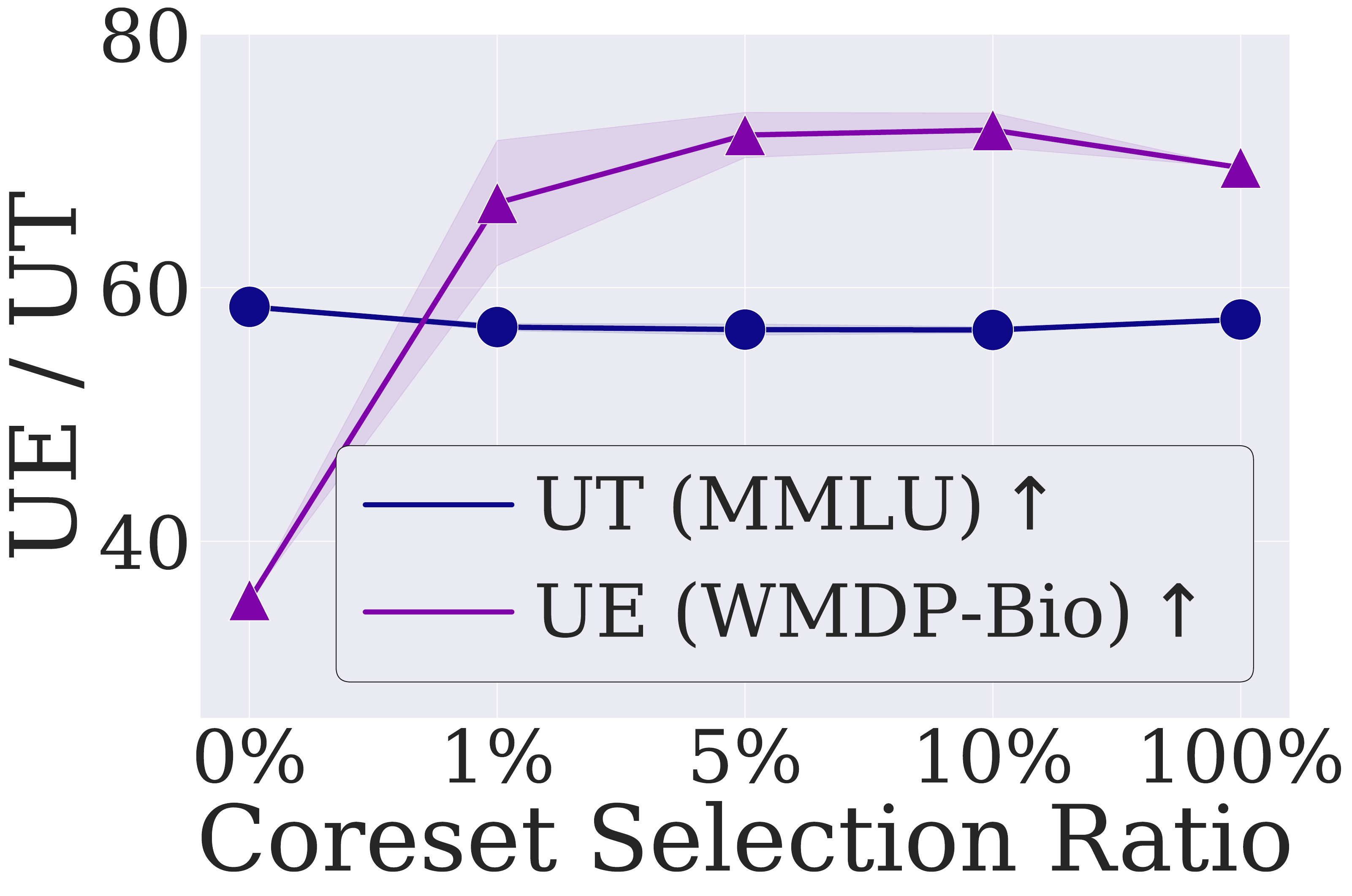}    &  
 \hspace*{-3.9mm}
 \includegraphics[width=.225\textwidth]{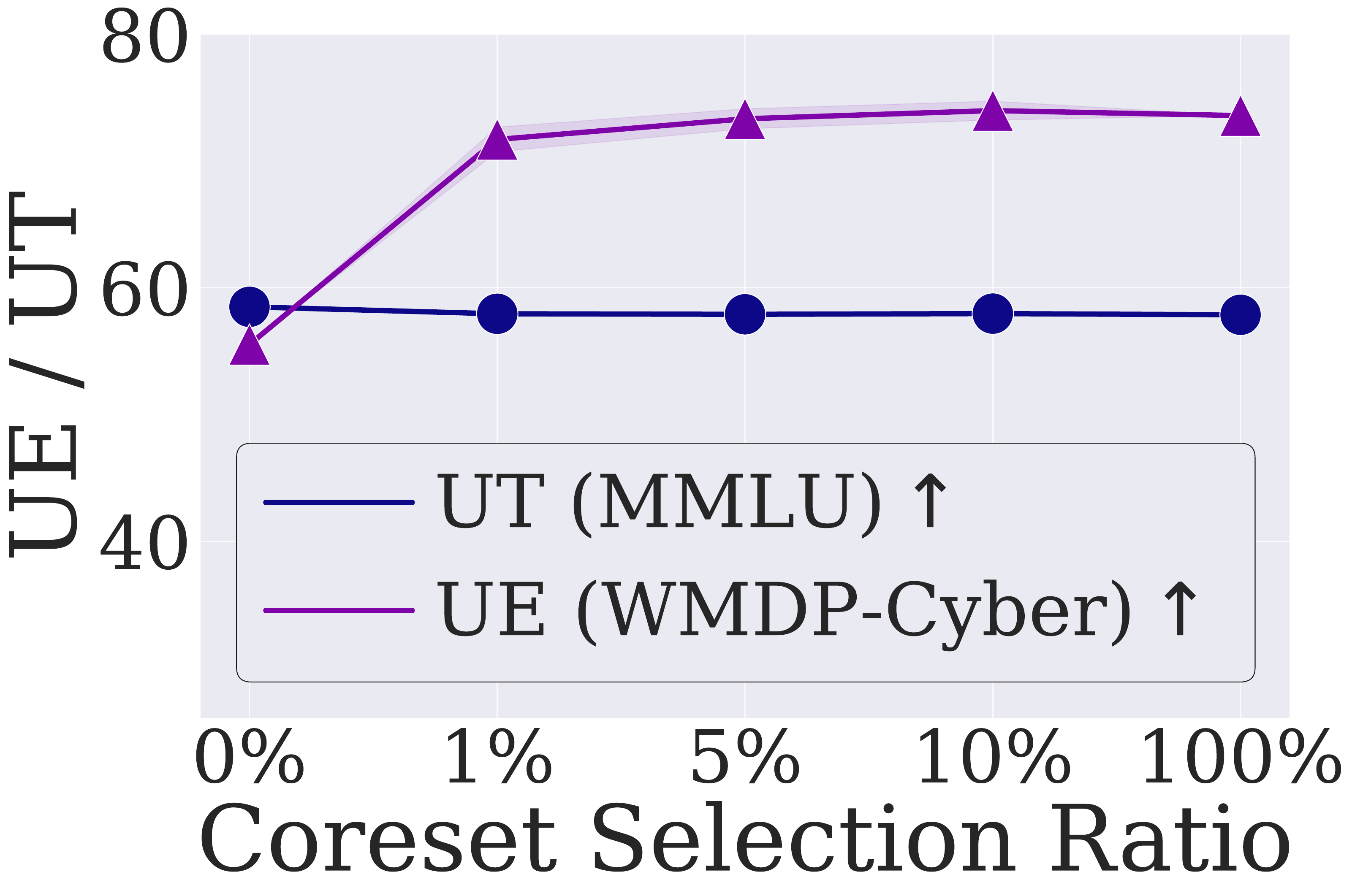} &
 \hspace*{-3.9mm}
 \includegraphics[width=.225\textwidth]{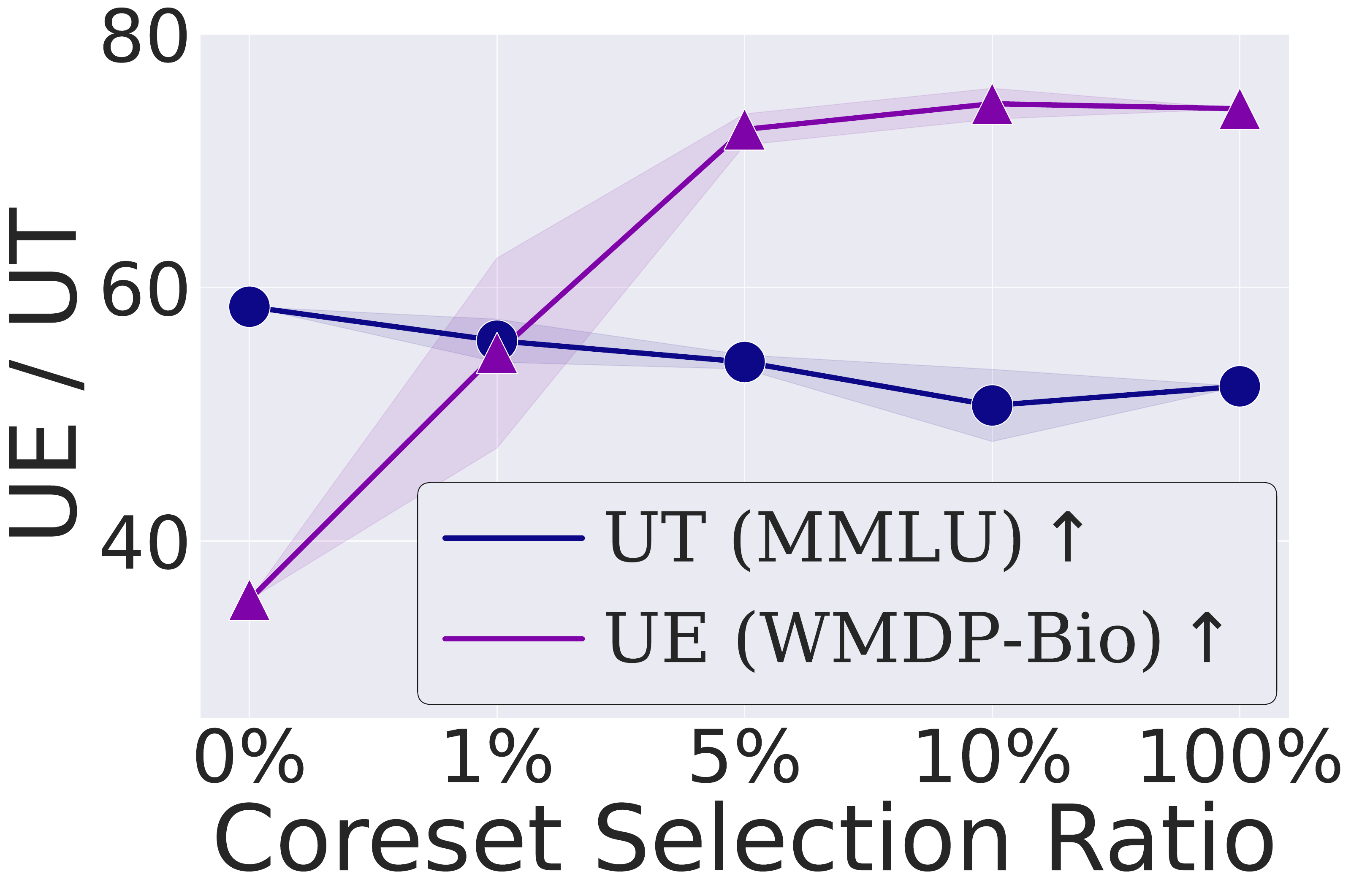}    &  
\hspace*{-3.9mm}
 \includegraphics[width=.225\textwidth]{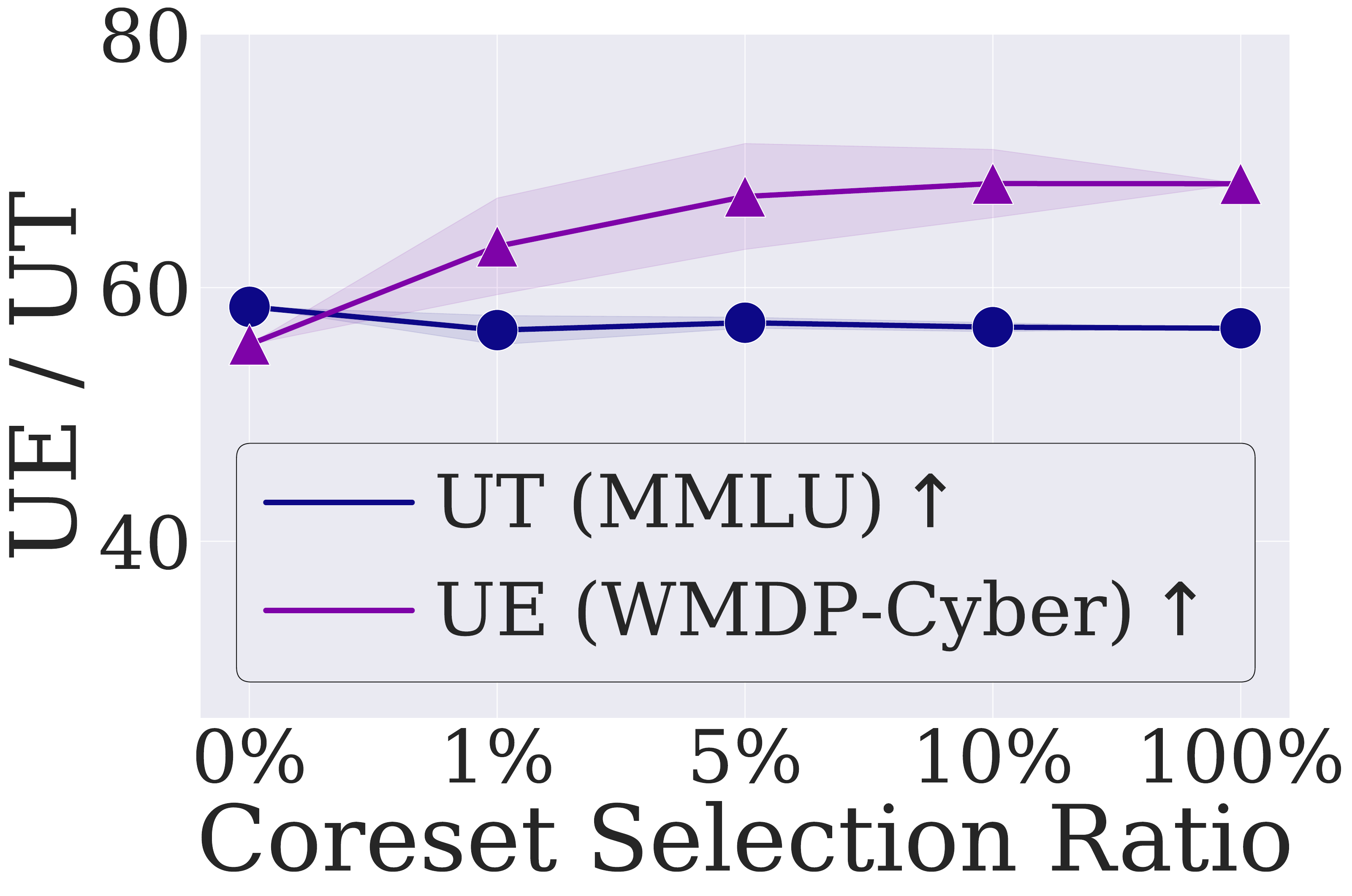} \\
 
\hspace*{-3mm}

{\footnotesize{(a) RMU, WMDP-Bio}}
 \hspace*{-3.9mm}
& {\footnotesize{(b) RMU, WMDP-Cyber}}
 \hspace*{-3.9mm}
& {\footnotesize{(c) NPO, WMDP-Bio}}  
 \hspace*{-3.9mm}
& {\footnotesize{(d) NPO, WMDP-Cyber}} \\

  \hspace*{-1mm}
  
\includegraphics[width=.225\textwidth]{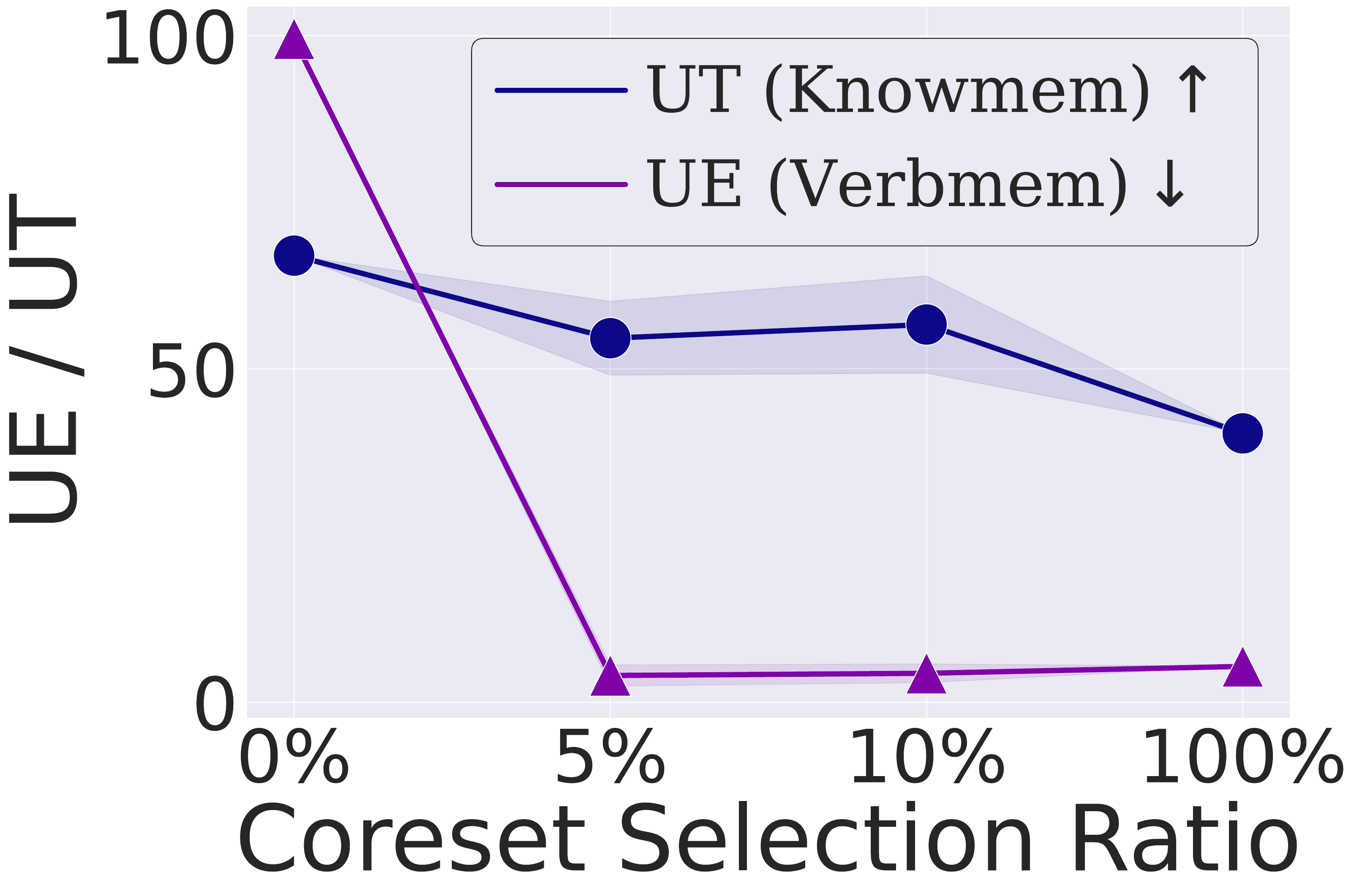}    &  
 \hspace*{-3.9mm}
 \includegraphics[width=.225\textwidth]{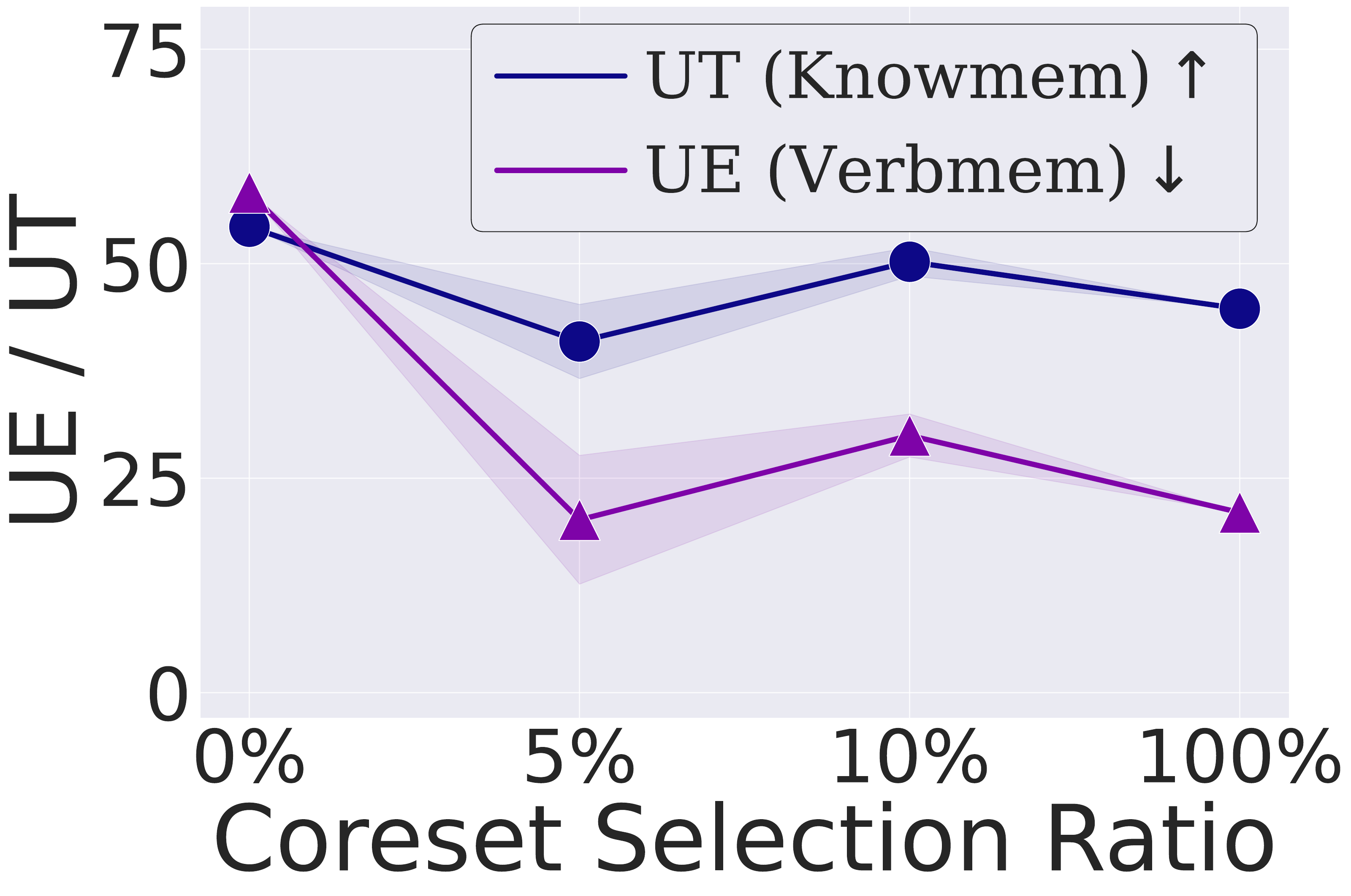} &
 \hspace*{-3.9mm}
 \includegraphics[width=.225\textwidth]{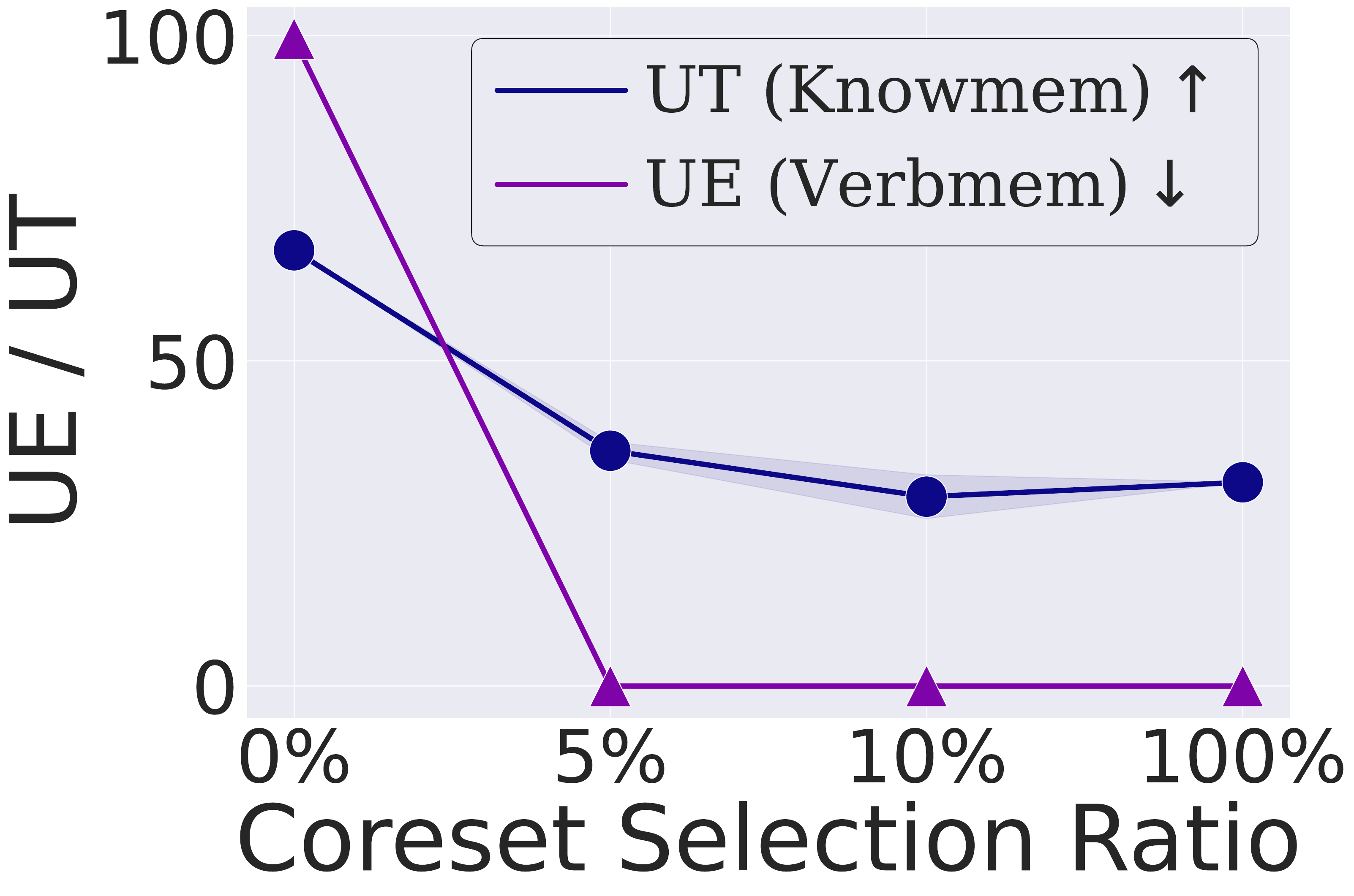}    &  
\hspace*{-3.9mm}
 \includegraphics[width=.225\textwidth]{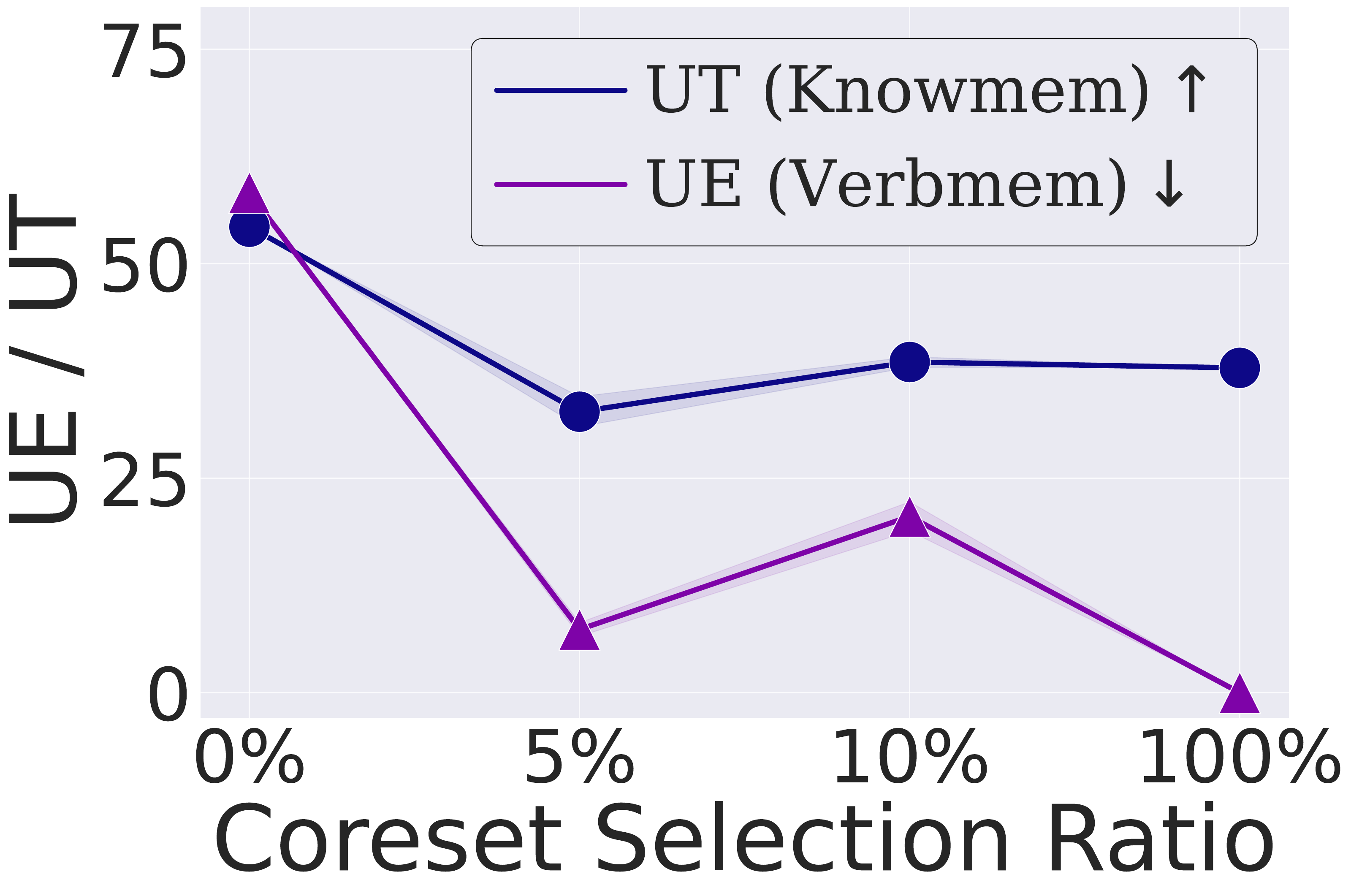} \\

\hspace*{-3mm}

{\footnotesize{(e) RMU, MUSE-Books}}
 \hspace*{-3.9mm}
& {\footnotesize{(f) RMU, MUSE-News}}
 \hspace*{-3.9mm}
& {\footnotesize{(g) NPO, MUSE-Books}}   \hspace*{-3.9mm}
 & {\footnotesize{(h) NPO, MUSE-News}} \\
\end{tabular}
\vspace*{-3mm}
\caption{\small{
Consistent \random{}-based coreset unlearning performance in terms of UT and UE across against the coreset selection ratio. The performance is averaged over 5 independent  trials 
for random coreset selection, with variance indicated by the shaded regions. 
(a)-(d) correspond to the results of applying a specific unlearning method (RMU or NPO) to a benchmark dataset 
(WMDP-Bio, WMDP-Cyber, MUSE-Books, or MUSE-News).
Following the benchmark setting,
unlearning is performed using Zephyr-7B-$\beta$ on WMDP,  LLaMA2-7B on MUSE-News, and ICLM-7B on MUSE-Books. 
}} 
\label{fig: random_sufficient}
\vspace*{-1mm}
\end{figure}

  \textbf{Fig.\,\ref{fig: random_sufficient}} shows the consistent \random{}-enabled coreset effect in LLM unlearning using RMU and NPO across four datasets: WMDP-Bio, WMDP-Cyber, MUSE-Books, and MUSE-News. The coreset selection ratio ranges from 0\% to 100\%, where 0\% corresponds to the original (pre-unlearning) model, and 100\% represents standard unlearning using the entire forget set. 
In Fig.\,\ref{fig: random_sufficient}(a)–(d), which shows WMDP-based unlearning, we observe that both RMU and NPO exhibit a clear coreset effect, evidenced by lossless UE and preserved UT as the coreset ratio increases  comparable to that achieved with the full forget set (100\%). Notably, RMU appears more effective in the low-data regime, achieving strong performance with as little as 1\% of the forget set, whereas NPO requires around 5\% to reach similar results in WMDP.
As shown for MUSE-based unlearning in Fig.\,\ref{fig: random_sufficient}(e)–(h), both RMU and NPO exhibit a consistent {coreset effect} in the context of {data unlearning}, as reflected by reductions in verbatim memorization. Notably, UT, measured by KnowMem, even improves when using a smaller coreset size; see Fig.\,\ref{fig: random_sufficient}(e).

Next, we compare the unlearning performance of \random{}-based coreset selection with three additional, more sophisticated coreset selection methods, \grand{} \citep{paul2021deep}, 
\moderate{} \citep{xia2022moderate}, and \mink{}  \citep{shi2024detecting}. The first two methods are classic coreset selection techniques adapted for LLMs, where \grand{} ranks forget samples based on the gradient norm of the unlearning objective in \eqref{eq: LLM_MU}, and 
\moderate{} clusters forget samples based on their associated deep representations, then ranks them by their distance to the cluster center.
\mink{} was originally developed to determine whether a given text appears in the original pretraining dataset, producing a data memorization score, where a higher value indicates stronger memorization of the data. In our setting, we select the unlearning coreset based on the top memorization scores.
We refer readers to Appendix\, \ref{app: coreset_selection} for details in the above coreset selection methods.

\begin{table}[htb]
\vspace{-5mm}
\begin{center}
\caption{\small{Coreset unlearning performance (UE and UT, consistent with Fig.\,\ref{fig: random_sufficient}) using RMU and NPO on WMDP-Bio and WMDP-Cyber, evaluated using Zephyr-7B-$\beta$ across varying coreset selection ratios (0\%, 5\%, 10\%, 100\%) and selection methods (\random{}, \grand{}, \moderate{}, \mink{}). Here 0\% and 100\% refer to the pre-unlearning case (w/o using any forget data) and the standard unlearning case (w/ the full forget set), respectively. The performance for \random{}-based selection is reported in the form $a \pm b$, where $a$ is the mean and $b$ is the standard deviation, computed over 5 independent trials. The performance of other data selection metrics are reported using the mean over 2 trials. 
}}
\vspace{3mm}
\label{tab: wmdp-bio}
\resizebox{\textwidth}{!}{
\begin{tabular}{c|c|c|c|c|c|c|c|c|c}

\toprule[1pt]
\midrule

\multicolumn{1}{c|}{\multirow{2.5}{*}{
\begin{tabular}{c}
   \textbf{Coreset}
   \\
    \textbf{Ratio} 
\end{tabular}
}} 
& 
\multicolumn{1}{c|}{\multirow{2.5}{*}{
\begin{tabular}{c}
   \textbf{Unlearning} \\
     \textbf{Method} 
\end{tabular}
}} 
& 
\multicolumn{2}{c|}{ \textbf{RMU} on \textbf{WMDP-Bio} } 
& 
\multicolumn{2}{c|}{\textbf{NPO} on \textbf{WMDP-Bio} } 
& 
\multicolumn{2}{c|}{\textbf{RMU} on \textbf{WMDP-Cyber} } 
& 
\multicolumn{2}{c}{\textbf{NPO} on \textbf{WMDP-Cyber} } \\
\cmidrule{3-10}

&& 
\multicolumn{1}{c|}{\textbf{UE} ($\uparrow$)} 
& 
\multicolumn{1}{c|}{\textbf{UT} (MMLU) ($\uparrow$)} 
& 
\multicolumn{1}{c|}{\textbf{UE} ($\uparrow$)}
& 
\multicolumn{1}{c|}{\textbf{UT} (MMLU) ($\uparrow$)} 
& 
\multicolumn{1}{c|}{\textbf{UE} ($\uparrow$)} 
& 
\multicolumn{1}{c|}{\textbf{UT} (MMLU) ($\uparrow$)} 
& 
\multicolumn{1}{c|}{\textbf{UE} ($\uparrow$)}
& 
\multicolumn{1}{c}{\textbf{UT} (MMLU) ($\uparrow$)} 
\\
\midrule
\multirow{1}{*}{0\%}
& No unlearning
& 35.35
& 58.48
& 35.35
& 58.48
& 55.51
& 58.48
& 55.51
& 58.48
 \\

\midrule
\multirow{1}{*}{100\%}
& Full forget set
& 69.46 
& 57.48 
& 74.11 
& 52.20 

& 73.57
& 57.85 
& 68.19 
&  56.79 
 \\

\midrule

\multirow{4}{*}{10\%} 
& \random{}  
& 72.43$_{\pm{1.34}}$
& 56.66$_{\pm{0.24}}$ 
& 74.50$_{\pm{1.22}}$
&  50.69$_{\pm{2.85}}$

& 73.96$_{\pm{0.74}}$
& 57.95$_{\pm{0.15}}$  
& 68.21$_{\pm{2.69}}$ 
&  56.88$_{\pm{0.35}}$ 
 \\
& \grand{}
& 71.30 
& 56.98 
& 71.54
&  52.40  
&73.43
& 57.25
&67.64
&56.71
 \\
& \moderate{} 
& 70.86
& 56.66 
&  75.92 
& 52.36 
& 73.53
& 57.20
& 69.65
& 56.69 
 \\
& \mink{} 
& 70.61
& 56.86
& 74.59
& 52.71
&73.91
&57.12
&69.19
&57.03
\\

\midrule
\multirow{4}{*}{5\% } 
& \random{} 
& 72.03$_{\pm{1.78}}$
& 56.69$_{\pm{0.44}}$ 
& 72.47$_{\pm{1.23}}$
&  54.12$_{\pm{0.56}}$

& 73.32$_{\pm{0.79}}$
& 57.89$_{\pm{0.12}}$ 
& 67.19$_{\pm{4.17}}$ 
&  57.23$_{\pm{0.43}}$
\\
& \grand{} 
& 72.59
& 56.78 
& 75.19 
&  56.55  
& 72.88
& 57.64
& 65.85
& 56.58
\\
& \moderate{} 
& 73.78 
& 56.90  
& 71.72 
& 56.18 
& 73.56
& 57.52
& 67.47
& 55.72
\\
& \mink{} 
& 69.48
& 57.29
& 70.91
& 55.67
& 72.84
& 57.58
& 69.15
& 57.19\\

\midrule
\bottomrule[1pt]
\end{tabular}
}
\vspace*{-3mm}
\end{center}
\end{table}


\reftab{tab: wmdp-bio}
presents the coreset unlearning performance across different coreset selection methods on WMDP.  
As we can see,
the coreset effect remains robust across different coreset selection methods.
Although random selection introduces  performance variance across trials due to differing coreset realizations, more sophisticated selection strategies typically offer only marginal improvements in UE or UT over {\random{}}, and these gains generally fall within the variance observed under {\random{}}.
Notably, compared to standard unlearning (\textit{i.e.}, using the full 100\% forget set), 5\% coreset unlearning can consistently achieve lossless UE and UT, demonstrating its effectiveness even in low-data regimes.
This consistent performance is also observed on the MUSE benchmark, as shown in \reftab{tab: muse}.

\noindent 
\textbf{Explaining the sufficiency of coreset unlearning: A keyword perspective.}
To explain the consistently strong coreset-like effect, we analyze the unlearning behavior of \random{}-based coreset selection from a keyword perspective.
Given a coreset, we extract its corresponding keyword set by leveraging LLM-as-a-judge (OpenAI o1 \citep{jaech2024openai}), guided by our proposed prompt (detailed in Appendix\,\ref{app: prompt}), to identify words most relevant to the targeted unlearning concept or knowledge (\textit{e.g.}, {biosecurity}).
We then perform unlearning on this keyword set to assess whether the majority of the unlearning efficacy can be attributed to this significantly smaller subset, in comparison to unlearning conducted on the full coreset (from which the keywords were extracted) and the original full forget set $\Df$.

\begin{table}[htb]
\begin{center}
\vspace*{-5.8mm}
\caption{\small{
Generation of different unlearned models for the sentence completion task using a part of the forget set of WMDP-Bio as input.
Keywords are marked in {light cyan} within the forget sample. The unlearning task and setting follow the same configuration as  in Fig.\,\ref{fig: random_sufficient}(a).
} 
}
\vspace*{-1mm}
\resizebox{\textwidth}{!}{
\begin{tabular}{c|c|c|c}
\toprule[1pt]
\midrule
\multicolumn{1}{c|}{\multirow{3}{*}{\textbf{Forget data sample from $\Df$ (WMDP-Bio) w/ extracted \colorbox{LightCyan}{keywords}}
}}  
& 
\multicolumn{1}{c|}{\multirow{3}{*}{
\begin{tabular}{c}
   \textbf{Full Forget Set ($\Df$)} \\
   \textbf{Unlearned Model}\\
    \textbf{Generation} 
\end{tabular}
}}
 & 
\multicolumn{1}{c|}{\multirow{3}{*}{
\begin{tabular}{c}
   \textbf{Coreset (5\% of $\Df$)} \\
   \textbf{Unlearned Model}\\
    \textbf{Generation} 
\end{tabular}
}}
& 
 \multicolumn{1}{c}{\multirow{3}{*}{
\begin{tabular}{c}
   \textbf{Keywords Unlearned}
   \\
    \textbf{Model Generation} 
\end{tabular}
}}
\\
&&&
\\
&&&\\
\midrule
 \multicolumn{1}{c}{\multirow{9}{*}{
\begin{tabular}{c}
The most common \colorbox{LightCyan}{pathogen} isolated from \colorbox{LightCyan}{urine cultures} is \colorbox{LightCyan}{Escherichia coli},\\ 80–90\%. However, other \colorbox{LightCyan}{bacteria} that were rarely isolated previously are\\ now rising (\colorbox{LightCyan}{Proteus}, \colorbox{LightCyan}{Citrobacter}, \colorbox{LightCyan}{Enterobacter}, and \colorbox{LightCyan}{Serratia} species). \colorbox{LightCyan}{E.coli} \\ can produce \colorbox{LightCyan}{extended-spectrum} \colorbox{LightCyan}{$\beta$-lactamase} (\colorbox{LightCyan}{ESBL}) \colorbox{LightCyan}{enzymes}, which provide\\
resistance against drugs like \colorbox{LightCyan}{penicillins}, \colorbox{LightCyan}{extended-spectrum} \colorbox{LightCyan}{cephalosporins},\\ and \colorbox{LightCyan}{monobactams}. These \colorbox{LightCyan}{ESBL-producing} \colorbox{LightCyan}{bacteria} are associated with \underline{\hspace*{5mm}} 
\end{tabular}
}}
&
\multicolumn{1}{c}{\multirow{9}{*}{
\begin{tabular}{c}
\textbackslash{n}\textbackslash{n}Many, there),"""v,\textbackslash{n}""B,s.\textbackslash{n}\\Where they\textbackslash{n}should\textbackslash{n}@(R)\\
Rory()\textbackslash{n}\textbackslash{n}\#a\textbackslash{n}where
\textbackslash{n}J) .\textbackslash{n}B\\
\textbackslash{n}(L.)\textbackslash{n}T\textbackslash{n}\textbackslash{n}R)Mart\textbackslash{n}\textbackslash{n}\#\#\#more\textbackslash{n}\textbackslash{n} \\ 
for\textbackslash{n}\textbackslash{n}\#(v)\textbackslash{n}\textbackslash{n}\textbackslash{n}\textbackslash{n}there\textbackslash{n}\textbackslash{n}None\\
\textbackslash{n}\textbackslash{n}to\textbackslash{n} \textbackslash{n}to\textbackslash{n}\textbackslash{n}
to\textbackslash{n}\textbackslash{n}where\textbackslash{n}\textbackslash{n}to\\
\textbackslash{n} 
\textbackslash{n}had\textbackslash{n}\textbackslash{n}to\textbackslash{n}\textbackslash{n}based\textbackslash{n}\textbackslash{n}

\end{tabular}
}}
&
\multicolumn{1}{c|}{\multirow{9}{*}{
\begin{tabular}{c}
--could to réve $\gamma$ \\
révère, $\rho$ $\beta$ B, \\
$\tau$ $\alpha$ x$\rho$$\epsilon$ \\
$\nu$$\tau$$\epsilon$$\sigma$$\beta$$\alpha$, \\
...
\end{tabular}
}}
& \multicolumn{1}{c}{\multirow{9}{*}{
\begin{tabular}{c}

'sometimes' were 'few' \\
in-levels.\textbackslash{n}\textbackslash{n}'The'\\
is a 'zero'-levels.\\
\textbackslash{n}\textbackslash{n}'The' is 'zero'-\\
'$\mu$'-'n'-'i'-'o'-'t'\\
-'100'-'n'-'x'-'i'-\\
...


\end{tabular}
}}
\\
&&&\\
&&&\\
&&&\\
&&&\\
&&&\\
&&&\\
&&&\\
&&&\\

\midrule
 \multicolumn{1}{c}{\multirow{11}{*}{
\begin{tabular}{c}
Since their first use as expression \colorbox{LightCyan}{vectors} in the 1980s, \colorbox{LightCyan}{Ad} \colorbox{LightCyan}{vectors} have received \\
tremendous attention as \colorbox{LightCyan}{gene} delivery vehicles for \colorbox{LightCyan}{vaccine} \colorbox{LightCyan}{antigens}. They have \\ been 
extensively tested as \colorbox{LightCyan}{vaccine} delivery systems in several pre-\colorbox{LightCyan}{clinical} and \\
\colorbox{LightCyan}{clinical}
studies for a number of \colorbox{LightCyan}{infectious} \colorbox{LightCyan}{diseases} including \colorbox{LightCyan}{measles}, \\
\colorbox{LightCyan}{hepatitis-B}, \colorbox{LightCyan}{rabies}, \colorbox{LightCyan}{anthrax}, \colorbox{LightCyan}{Ebola}, severe \colorbox{LightCyan}{acute} \colorbox{LightCyan}{respiratory} \colorbox{LightCyan}{syndrome} \\
(\colorbox{LightCyan}{SARS}), human \colorbox{LightCyan}{immunodeficiency} \colorbox{LightCyan}{virus} 1 (\colorbox{LightCyan}{HIV-1}), \colorbox{LightCyan}{malaria}, \colorbox{LightCyan}{tuberculosis}, \\
and \colorbox{LightCyan}{influenza}. There are two basic types of \colorbox{LightCyan}{Ad} \colorbox{LightCyan}{vectors} that are being \\
utilized for 
\colorbox{LightCyan}{gene} delivery applications. The first type of \colorbox{LightCyan}{Ad} \colorbox{LightCyan}{vectors},\underline{\hspace*{5mm}} 
\end{tabular}
}}
&
\multicolumn{1}{c}{\multirow{11}{*}{
\begin{tabular}{c}

.@the...\textbackslash{n}\textbackslash{n}\#\#\#.more\textbackslash{n}\textbackslash{n}.\textbackslash{n}\textbackslash{n}\#\#\#.s\textbackslash{n}\\
vin.x 
\textbackslash{n}.\textbackslash{n}\textbackslash{n}\#\#,\textbackslash{n}inway.\textbackslash{n}x.\textbackslash{n}\\
\textbackslash{n}@\textbackslash{n}@@
in.\textbackslash{n}
\#\#\#\textbackslash{n}xer\textbackslash{n}more\textbackslash{n}\textbackslash{n}b\textbackslash{n}\\
no\textbackslash{n}\textbackslash{n}w
\textbackslash{n}\textbackslash{n}There\textbackslash{n}@x\textbackslash{n}h\textbackslash{n}\textbackslash{n}\\
no\textbackslash{n}\textbackslash{n}
scundo\textbackslash{n}there.\textbackslash{n}\textbackslash{n}no\textbackslash{n}\textbackslash{n}\\
how.\textbackslash{n}\textbackslash{n}
Cural.\textbackslash{n}sair\textbackslash{n}\textbackslash{n}hg\textbackslash{n}.\textbackslash{n}'

\end{tabular}
}}
&
\multicolumn{1}{c|}{\multirow{11}{*}{
\begin{tabular}{c}
.\textbackslash{n}\textbackslash{n}(none)\textbackslash{n}\textbackslash{n}.there..\\
iron.
törko.you.unde.undef.\\
und.
undund@.sUnd.none\textbackslash{n}\\
\textbackslash{n}undo
.@.every.all.they.just\textbackslash{n}\\
\textbackslash{n}W.x.wereunder.undo.yer\\
.@.\textbackslash{a}l.@x.@\textbackslash{n}\textbackslash{n}rex.undund.\\
universo@.under.all\\
\textbackslash{n}.there.under.rem@.unpret'
\end{tabular}
}}
& \multicolumn{1}{c}{\multirow{11}{*}{
\begin{tabular}{c}
who knew, were, \\
and, and, were,  \\
and, and, were, \\
were, and, were,\\
were, were, and, \\
were, were, were,\\
were, were, were,\\
...
\end{tabular}
}}
\\
&&&\\
&&&\\
&&&\\
&&&\\
&&&\\
&&&\\
&&&\\
&&&\\
&&&\\
&&&\\

\midrule
\bottomrule
\end{tabular}
}
\label{tab: keyword}
\vspace*{-2mm}
\end{center}

\end{table}

\textbf{Table\,\ref{tab: keyword}}  presents representative keyword examples and compares the generation of unlearned models across three different forget sets: the full $\Df$, a 5\% coreset, and a keyword-only subset extracted from the coreset.
The keyword set is constructed by filtering out non-keyword tokens and preserving the original order, as highlighted in the first column. As we can see, unlearning with just the keyword set remains highly effective, comparable to the full and coreset-based unlearning, as the model fails to recall the target knowledge and instead generates nonsensical responses to forget queries.


\begin{wrapfigure}{r}{50mm}
\vspace*{-3mm}
\centerline{
\begin{tabular}{c}
\hspace*{0mm}\includegraphics[width=.3\textwidth,height=!]{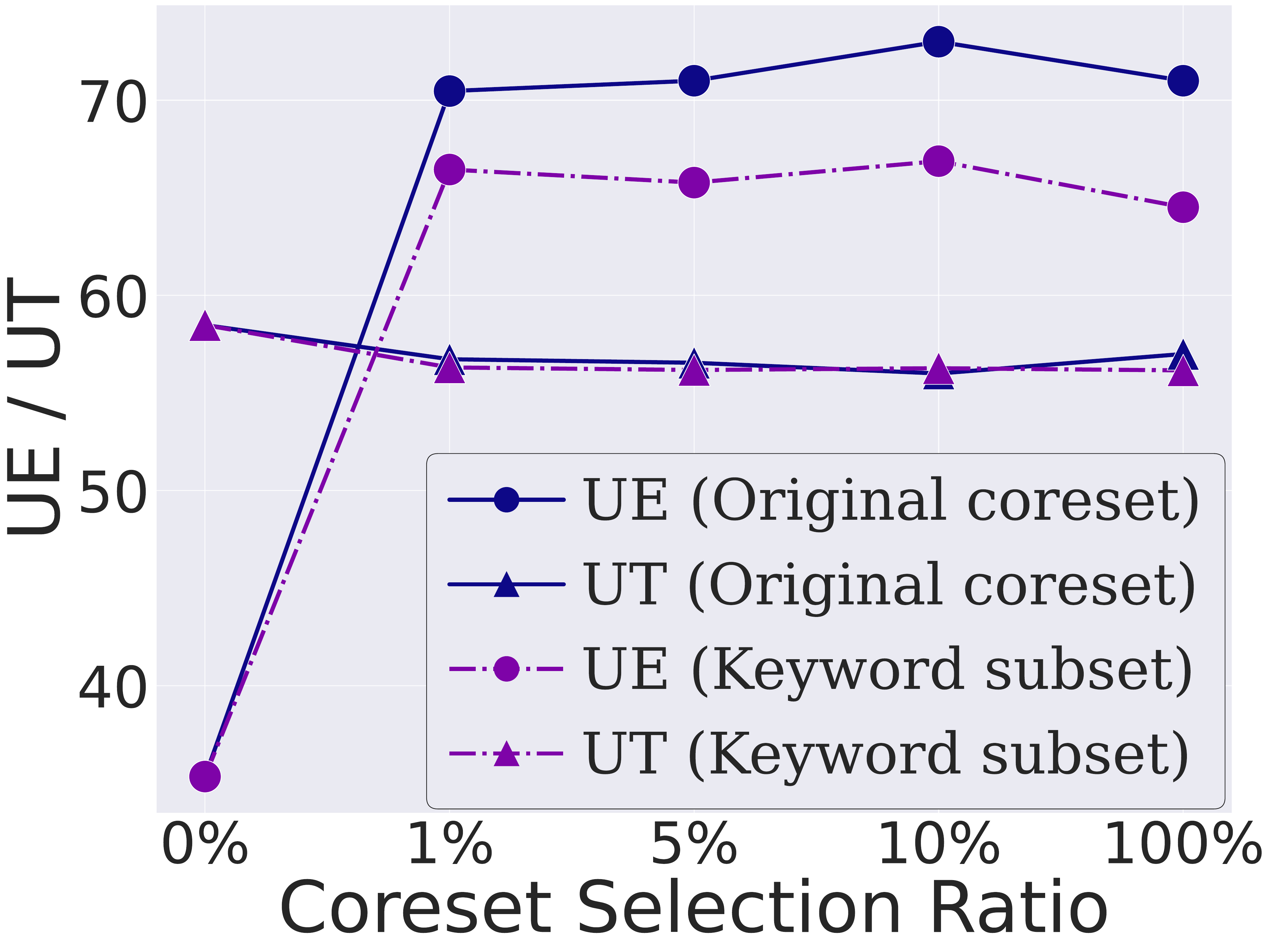}  
\end{tabular}}
\vspace*{-3mm}
\caption{\small{
Unlearning performance (UE and UT) using the original coreset and its keyword subset across varying coreset selection ratios for RMU-based unlearning on (WMDP-Bio, Zephyr-7B-$\beta$).
}
}
  \label{fig: keyword_unlearn}
  \vspace*{-3mm}
\end{wrapfigure}
For a thorough quantitative evaluation, \reffig{fig: keyword_unlearn} compares the unlearning performance (in both UE and UT) of using the keyword-only coreset and the original coreset against  the  coreset selection ratio on WMDP-Bio, following the same unlearning task setting as in Fig.\,\ref{fig: random_sufficient}(a). We perform 300 epochs of unlearning for the keyword-only forget sets.
As observed, although unlearning with the keyword-only forget sets does not fully match the UE of the full coreset, it captures the {majority of the unlearning effect} compared to the pre-unlearning baseline at 0\% coreset selection ratio. The contribution of using keyword unlearning is consistent against the coreset selection ratio. 
This suggests that the coreset effect may be driven by a small number of high-impact keywords within the coreset, which are sufficient to account for most of the unlearning performance. This, in turn, points to a surprising simplicity in the current unlearning benchmark tasks or their evaluation protocols.  Additional analysis on keyword overlap between the coreset and full forget set is provided in Appendix\,\ref{app: keyword_overlap}.

\vspace*{-3mm}
\section{On the Faithfulness of LLM Unlearning using Coresets}
\vspace*{-3mm}
In this section, we investigate the quality and faithfulness of coreset-based LLM unlearning from additional perspectives: 
(1) mode connectivity, which reflects the similarity of the unlearning loss landscape to that of full forget set-based unlearning; (2) the robustness of coreset-unlearned models against both input-based jailbreaking attacks and weight-based model fine-tuning post unlearning; and (3) additional utilities of unlearned LLMs beyond those captured by existing unlearning benchmarks.  
Unless otherwise specified, in this section we focus on LLM unlearning on WMDP using \random{}-based coreset selection, following the settings used in Sec.\,\ref{sec: consistency}.

\begin{figure}[htb]
\vspace*{-2mm}
\centering
\begin{tabular}{cccc}
  \hspace*{-2mm}
 \includegraphics[width=.23\textwidth]{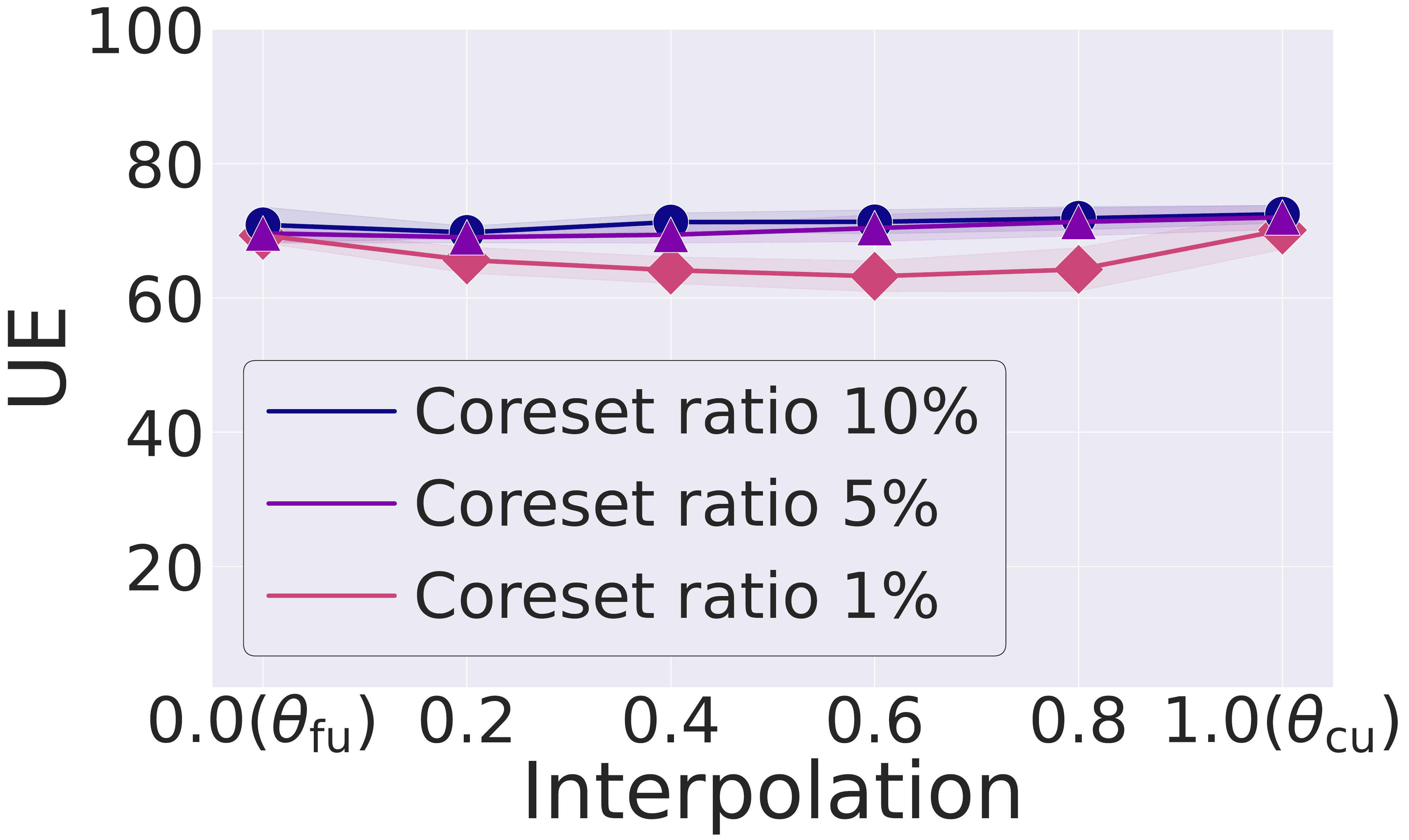}    &  
 \hspace*{-3.9mm}
 \includegraphics[width=.23\textwidth]{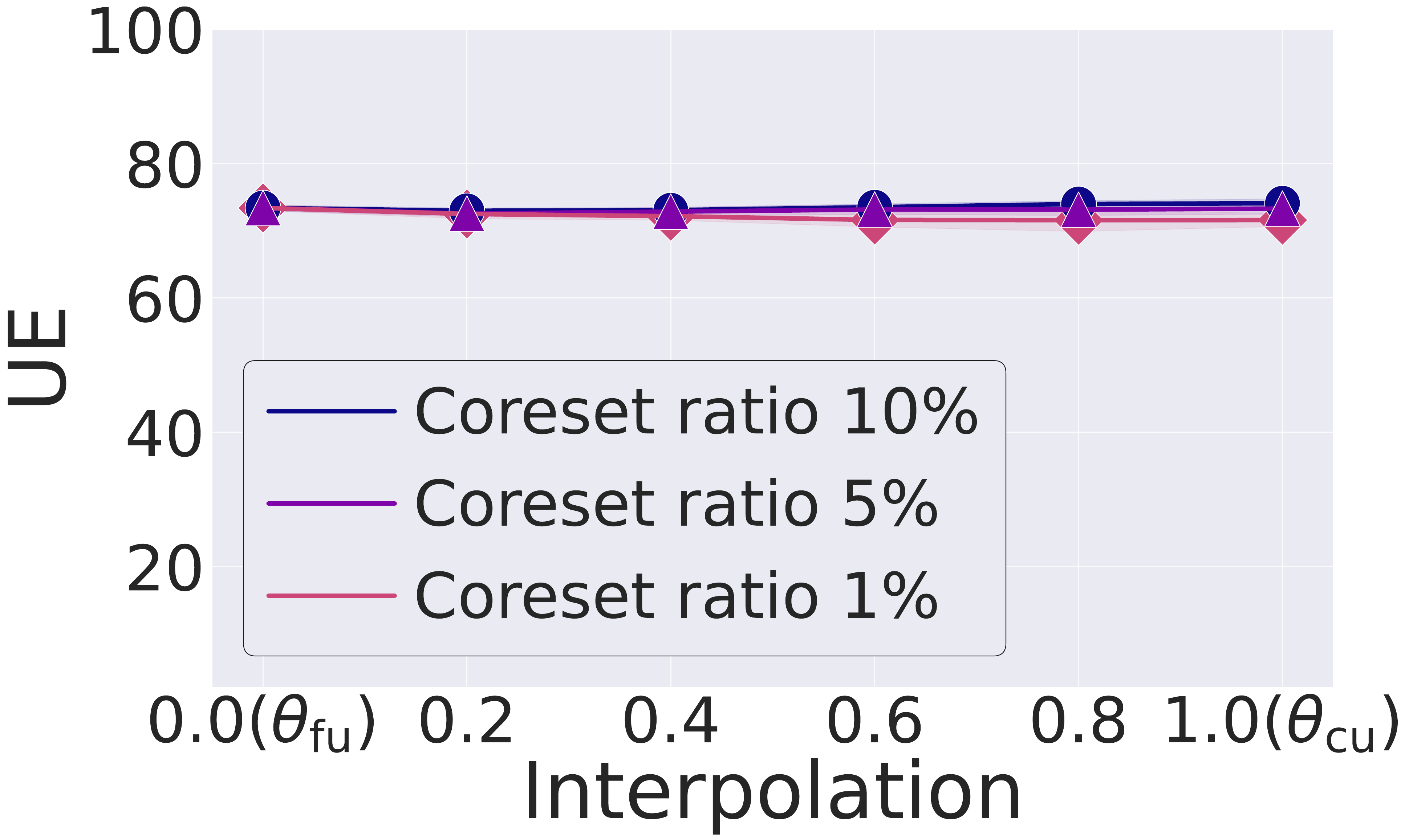} &
 \hspace*{-3.9mm}
 \includegraphics[width=.23\textwidth]{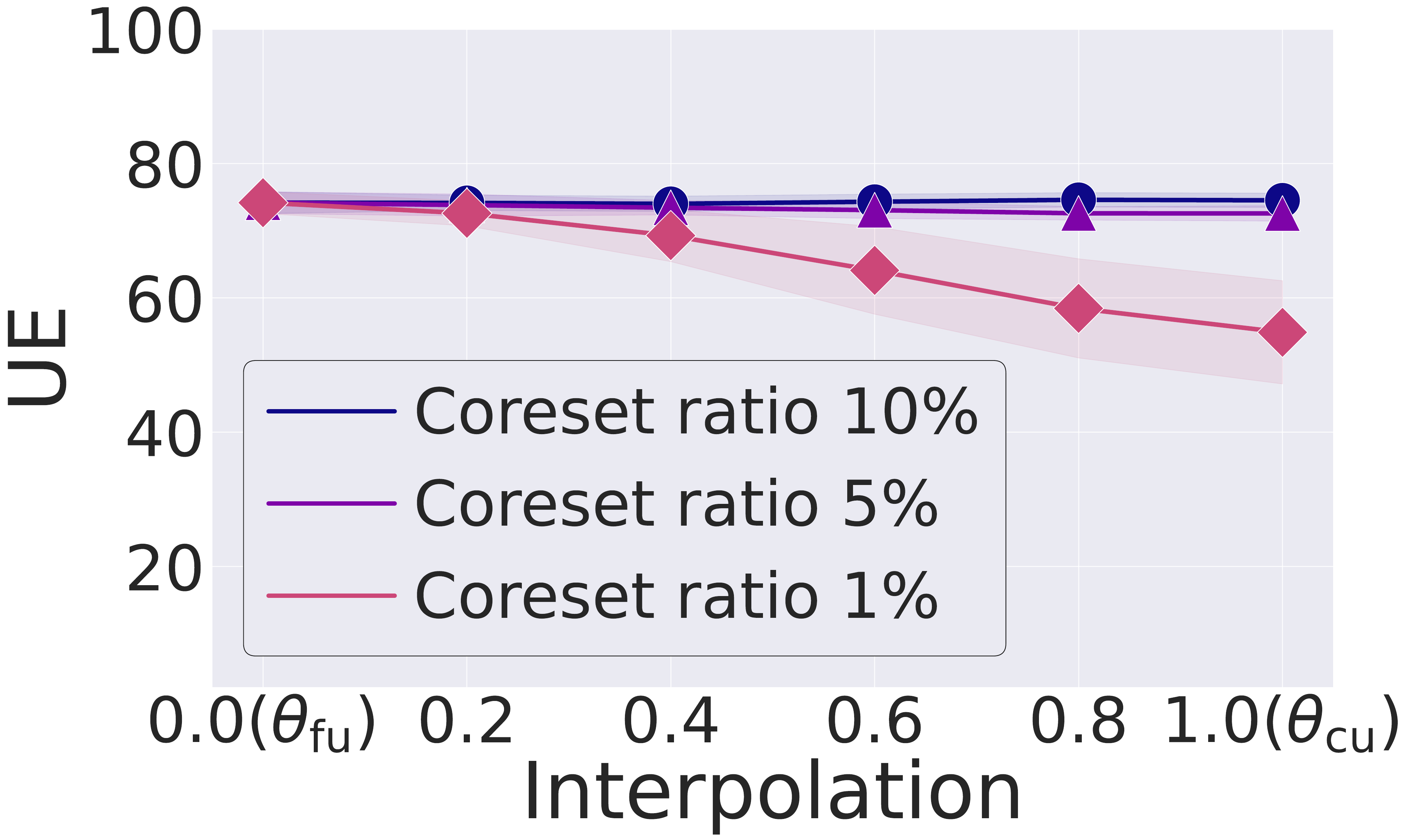}  
 &
 \hspace*{-3.9mm}
 \includegraphics[width=.23\textwidth]{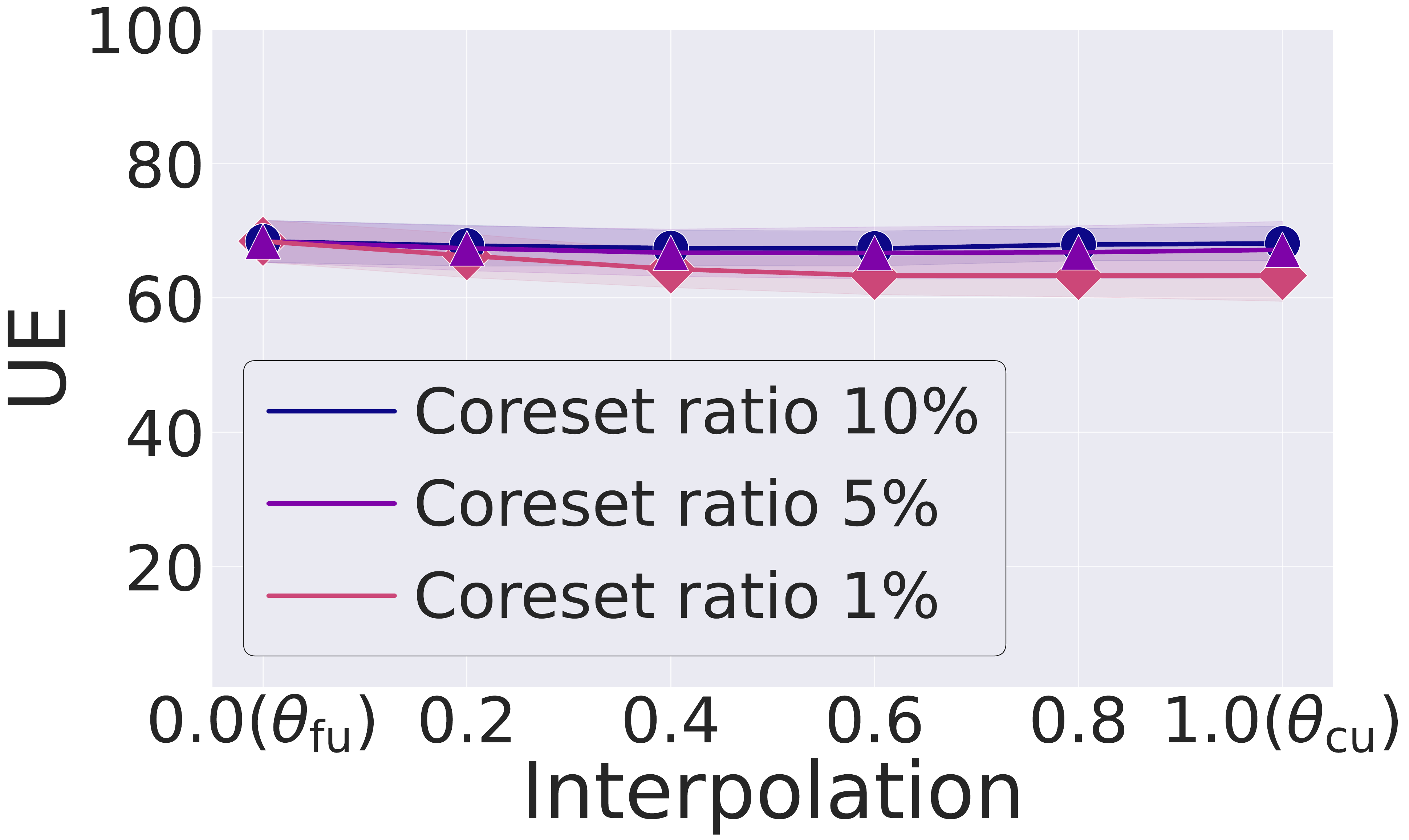}  
 \\
   \hspace*{-4mm}
{\footnotesize{(a) RMU, WMDP-Bio}}
&  \hspace*{-3.9mm}
{\footnotesize{(b)  RMU, WMDP-Cyber}}
&  \hspace*{-3.6mm}
{\footnotesize{(c) NPO, WMDP-Bio}}
&  \hspace*{-3mm}
{\footnotesize{(d) NPO, WMDP-Cyber}}
\\ 
\\
\end{tabular}
\vspace*{-6mm}
\caption{\small{LMC holds between coreset-unlearned model ($\btheta_\mathrm{cu}$) and the full forget set-unlearned model ($\btheta_\mathrm{fu}$), as evidenced by  UE against the interpolation coefficient $\alpha$  (x-axis). 
Here the coreset-unlearned models are obtained using \random{}-based coresets with the same setting 
as  in Fig.\,\ref{fig: random_sufficient}(a-d).
}}
  \label{fig:  modeconnection}
\end{figure}

\noindent 
\textbf{Mode connectivity: Coreset unlearning is as good as full forget set unlearning.}
Mode connectivity refers to the phenomenon where the minima found by two optimized ML models are connected by a path along which the model error does not increase, suggesting that the models reside in a shared or smoothly connected region of the loss landscape \citep{draxler2018essentially,freeman2016topology,qin2022exploring,garipov2018loss}. A strong form of connectivity is known as linear mode connectivity (\textbf{LMC}), where the interpolation path between two models is constrained to be linear in parameter space \citep{frankle2020linear}.
Therefore, we leverage LMC to investigate the similarity or potential discrepancy between a \underline{c}oreset-\underline{u}nlearned model ($\btheta_\mathrm{cu}$) and the \underline{f}ull forget set-\underline{u}nlearned model ($\btheta_\mathrm{fu}$).
\textbf{LMC for coreset unlearning is said to hold} if the unlearning evaluation (\textit{e.g.}, via UE) of the \textit{linearly interpolated model}
 $\btheta(\alpha) \Def ( \alpha \btheta_\mathrm{cu} + (1-\alpha) \btheta_\mathrm{fu} )$ remains approximately consistent as the interpolation coefficient   $\alpha \in [0, 1]$ varies, where $\btheta(1) = \btheta_\mathrm{cu}$ and $\btheta(0) = \btheta_\mathrm{fu}$ yield the endpoints of the linear path.  
\reffig{fig: modeconnection} illustrates   UE   of the interpolated model $\btheta(\alpha)$ against the interpolation coefficient $\alpha$  at forget coreset selection ratios of 1\%, 5\%, and 10\%.
As we can see,  the UE remains \textit{approximately constant} along the linear interpolation path between $\btheta_\mathrm{cu}$ and $\btheta_\mathrm{fu}$, indicating a \textit{strong} LMC between the two models (nearly perfect connectivity for 5\% and 10\% coresets). This provides strong evidence that the coreset-unlearned model resides in the same optimal basin as the full forget set-unlearned model, supporting the faithfulness of unlearning achieved via coresets.

\begin{wraptable}{r}{70mm}
\vspace*{-6mm}
\caption{\small{
Robustness of models unlearned using \random{} coreset selection on WMDP-Bio using RMU under Zephyr-7B-$\beta$, following the setting in Fig.\,\ref{fig: random_sufficient}(a).
Robustness is measured using the UE reduction after enhanced GCG attack. 
} 
}
\resizebox{0.5\textwidth}{!}{
\begin{tabular}{c|c|c|c}
\toprule[1pt]
\multicolumn{1}{c|}{\multirow{2.5}{*}{
\begin{tabular}{c}
     \textbf{Coreset} \\
     \textbf{Ratio}
\end{tabular}
}} 
& \multicolumn{2}{c|}{\textbf{UE}} 
& \multirow{1}{*}{\textbf{UE reduction}} \\
\cmidrule{2-4}
 & \multicolumn{1}{c|}{\textbf{Before Attack}} 
 & \multicolumn{1}{c|}{\textbf{After Attack}}
 & \multirow{1}{*}{\textbf{After Attack}} \\

\midrule
 100\%   
& 69.46 
& 47.71 
& 21.75\\

10 \%   
& 72.43$_{\pm{1.34}}$ 
& 53.39$_{\pm{0.02}}$   
& 19.04 \\

5 \%   
& 72.03$_{\pm{1.78}}$
& 51.29$_{\pm{0.03}}$   
& 20.74\\

\bottomrule
\end{tabular}
}
\label{tab: gcg}
\vspace*{-2mm}
\end{wraptable}


%
\noindent \textbf{Adversarial robustness against jailbreaking attacks.}
The lack of robustness in LLM unlearning has been highlighted by their vulnerability to jailbreaking attacks \citep{lucki2024adversarial, lynch2024eight, patil2023can}. 
Therefore, we investigate whether coreset-based unlearning introduces additional robustness limitations compared to unlearning with the full forget set.
\reftab{tab: gcg} presents the UE of coreset-unlearned models, alongside the full forget set-unlearned model (\textit{i.e.}, coreset ratio 100\%), under input-level jailbreaking attacks using the enhanced-GCG method \citep{lucki2024adversarial}, which generates adversarial prompt prefixes to elicit forgotten information. A lower UE after the attack indicates a more successful recovery of forgotten content. As shown, the coreset-unlearned models experience a similar degree of UE reduction as the full-set counterpart, demonstrating comparable robustness. These results further support the faithfulness and reliability of coreset-based unlearning.

\begin{wrapfigure}{r}{79mm}
\vspace*{-2mm}
\centering
\begin{tabular}{cc}
 \includegraphics[width=.23\textwidth]{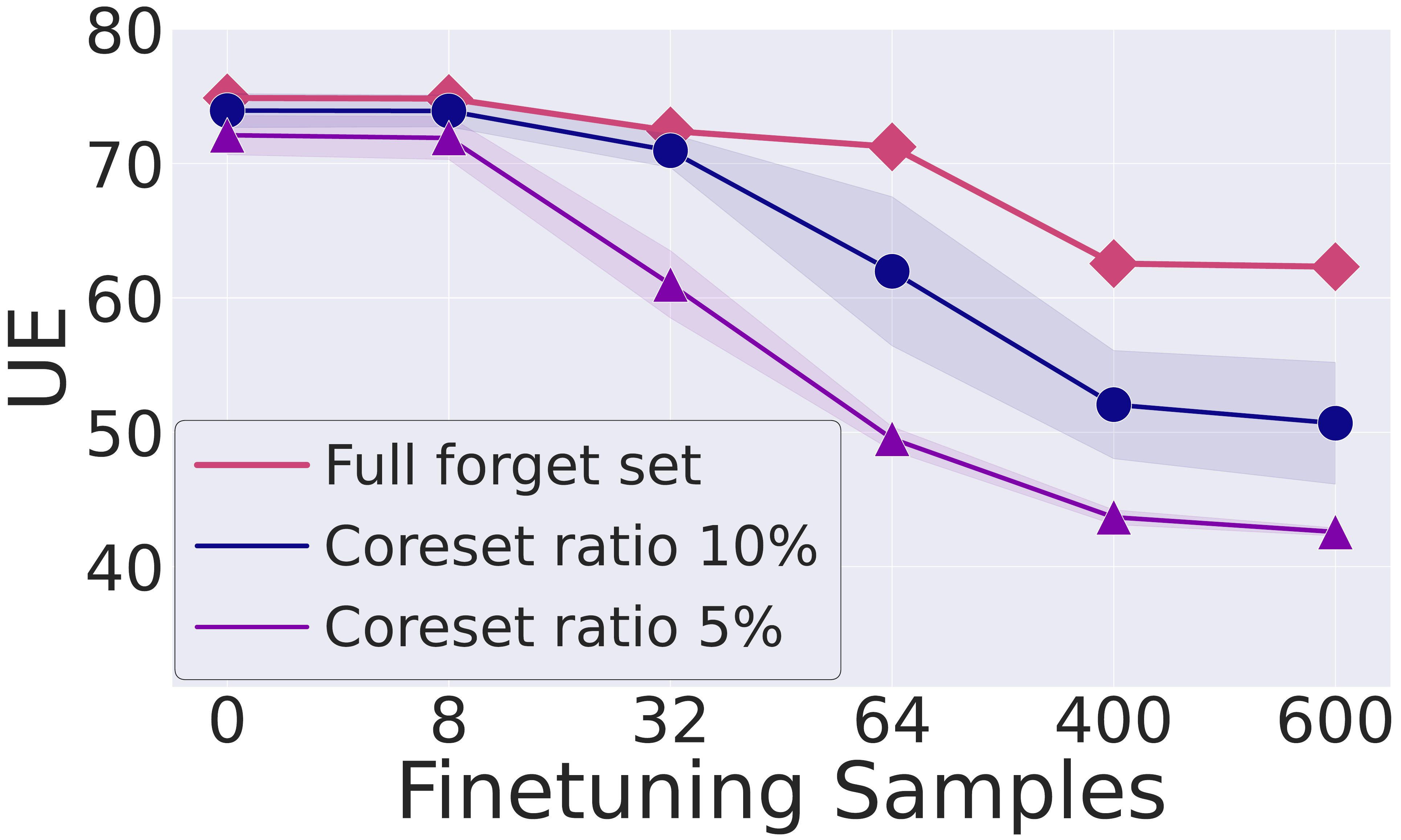}    &  
 \hspace*{-4mm}
 \includegraphics[width=.23\textwidth]{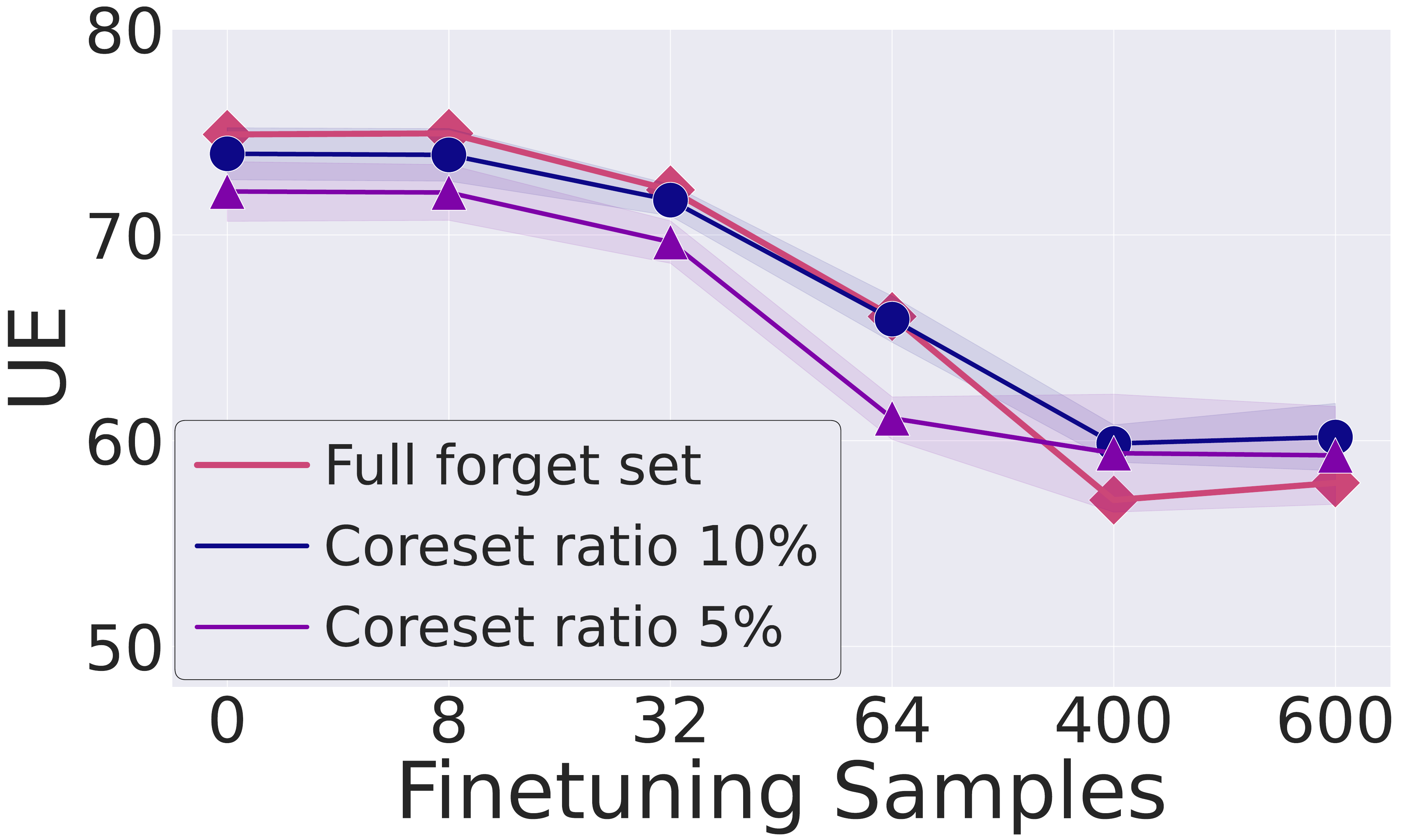} 
 \\
{\footnotesize{(a) GSM8k, WMDP-Bio}}
 \hspace*{-4mm}
& {\footnotesize{(b) AGNews, WMDP-Bio}}
\\
\includegraphics[width=.23\textwidth]{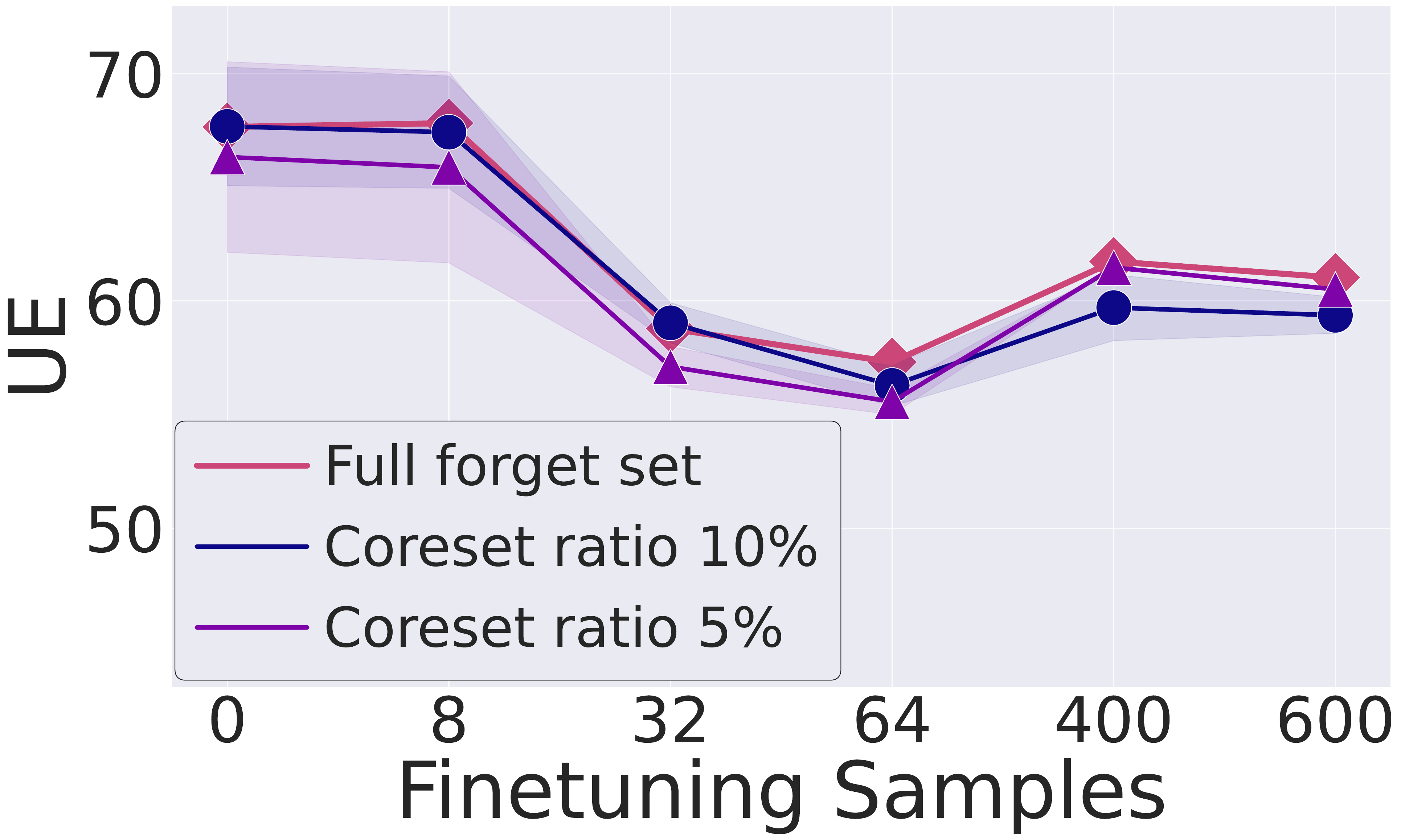}    &  
 \hspace*{-4mm}
 \includegraphics[width=.23\textwidth]{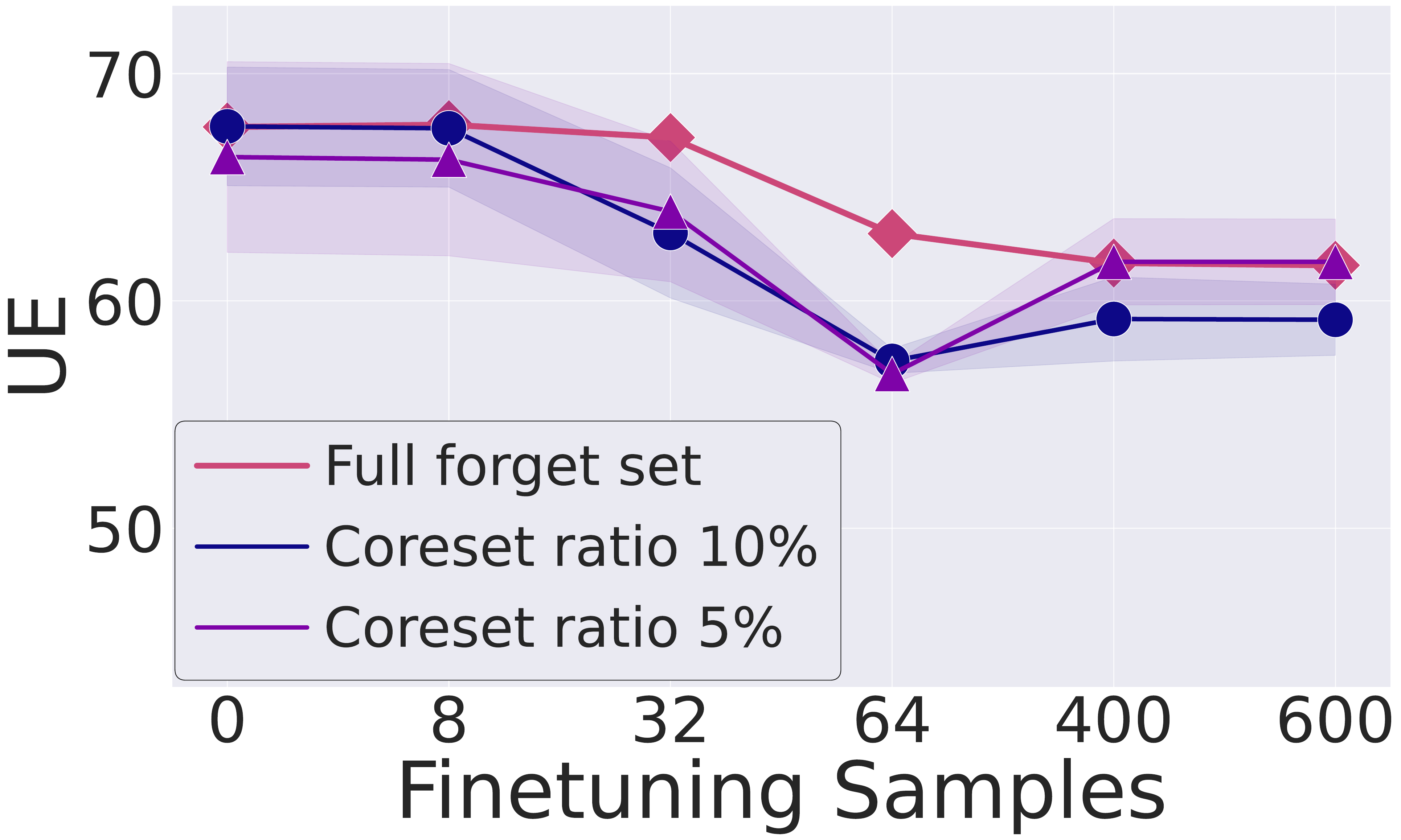} 
  \\
{\footnotesize{(d) GSM8k, WMDP-Cyber}}
 \hspace*{-4mm}
& {\footnotesize{(e) AGNews, WMDP-Cyber}}
\\
\end{tabular}
\vspace*{-3mm}
\caption{\small{Unlearning performance (UE) of \random{}-coreset unlearned models (using NPO  under  Zephyr-7B-$\beta$) against the number of fine-tuning samples. (a)-(f) Relearning using finetuning datasets (GSM8k, AGNews) for models unlearned on WMDP-Bio or WMDP-Cyber. The performance is averaged over 3 independent trials.
}}
  \label{fig: finetune}
\vspace*{-3mm}
\end{wrapfigure}
\noindent \textbf{Robustness against downstream fine-tuning.
}
Beyond
jailbreaking attacks, unlearning robustness has also been studied in 
model
fine-tuning, where forgotten knowledge can resurface
in unrelated downstream tasks \citep{hu2024jogging,deeb2024unlearning,lo2024large},
similar to such vulnerability of safety-aligned LLMs
\citep{qi2023fine}.
We finetune models unlearned on WMDP using NPO using GSM8k \citep{cobbe2021training} and AGNews \citep{zhang2015character} 
using 600 samples from the finetuning dataset. 
In \reffig{fig: finetune}, we demonstrate UE of unlearned models subjected to such finetuning. As expected, the models exhibit 'relearning' of unlearned information. 
However, in certain scenarios, such as  WMDP-Bio unlearned models finetuned using GSM8k in Fig.\,\ref{fig: finetune}(a), the speed and extent of relearning can clearly distinguish coreset-unlearned models from those unlearned using the full forget set.
These results highlight a potential downside of coreset unlearning: \textit{using a smaller forget set may compromise robustness to downstream fine-tuning compared to full-set unlearning.} This is likely because, when the relearn set is larger than  coreset, the influence of relearning can overpower the unlearning.



In Appendix\,\ref{app: utility},  we also show  that coreset-unlearned models retain utility on auxiliary tasks such as math addition/subtraction and TruthfulQA.


\vspace*{-3mm}
\section{Conclusion and  Discussion}
\vspace*{-3mm}

We establish a novel coreset effect in LLM unlearning, where an extremely small subset of the forget set can achieve lossless unlearning. This coreset effect particularly emerges when conditioned on sufficient unlearning training.
And 
this effect is shown to hold consistently across benchmarks (WMDP, MUSE) and unlearning methods (RMU, NPO). We further explain and validate this phenomenon from multiple perspectives, including keyword analysis, mode connectivity, and robustness.
While we establish this effect, we observe that many classic heuristic-based coreset selection methods do not outperform simple \random{} selection. 
Therefore, determining the `optimality' of random selection or identifying optimized, improved coresets remains a critical direction for future research.
Our keyword analysis also highlights the need for deeper investigation into the underlying mechanisms of unlearning, potentially through more advanced interpretability techniques. In other words, unlearning could be driven by salient tokens within the forget data, rather than relying on the entire forget dataset.
Additionally, the strong coreset effect observed in this work suggests potential `redundancy' within the forget sets used in current benchmarks. We hypothesize that this effect arises from inherent `correlations' among forget data points. If the forget examples were independently and identically distributed, the coreset effect might diminish significantly. This correlation-driven redundancy could make unlearning tasks \textit{easier than expected} under current evaluation protocols. 
We advocate for future benchmarks to incorporate the influence of forget dataset size and richer evaluation methods to better reflect realistic unlearning challenges.





\section*{Ethics Statement}


The ethical implications of our work on identifying the coreset effect and developing coreset-based unlearning methods lie in promoting responsible data handling and the safe use of large language models (LLMs). Our research advances machine unlearning techniques aimed at strengthening privacy protections, reducing sociotechnical harms, and  ensuring more controlled, trustworthy generative AI. 
At the same time, we recognize that the societal impact of unlearning is nuanced and continually evolving. This underscores the importance of ongoing reflection, transparent evaluation, and active engagement with the broader discourse on AI ethics and governance.
On the other hand, it is equally important to prevent the misuse of unlearning techniques to erase safe, beneficial, or ethically important information from LLMs, which could compromise model integrity and safety.



\section*{Acknowledgments} 
We thank the U.S. Department of Energy via Lawrence Livermore National Laboratory (LLNL) under Contract DE-AC52-07NA27344 and the LLNL LDRD Program under Project No. 23-ER-030 for their support (LLNL-JRNL-2004686). Soumyadeep Pal, Changsheng Wang,  and Sijia Liu were also partially supported by the National Science Foundation (NSF) CISE Core Program Award (IIS-2207052), the NSF CAREER Award (IIS-2338068), the ARO Award (W911NF2310343), and the  Amazon Research Award for AI in Information Security.


\bibliography{refs/MU,refs/Coreset,refs/models}
\bibliographystyle{colm2025_conference}

\clearpage

\input{secs/appendix}

\end{document}